\documentclass{article}
\usepackage{arxiv}

\usepackage[utf8]{inputenc} 
\usepackage[T1]{fontenc}    
\usepackage{hyperref}       
\usepackage{url}            
\usepackage{booktabs}       
\usepackage{amsfonts}       
\usepackage{nicefrac}       
\usepackage{microtype}      
\usepackage{lipsum}
\usepackage{algorithm}
\usepackage{algorithmic}
\usepackage{graphicx}
\usepackage{graphicx}
\usepackage{amsmath}
\usepackage{amssymb}
\usepackage{booktabs}
\usepackage{algorithm}
\usepackage{algorithmic}
\usepackage{float}

\usepackage{subfigure}
\usepackage{caption}
\usepackage{subcaption}

\usepackage{xcolor}
\usepackage{todonotes}
\usepackage{paralist}
\usepackage{multirow}
\usepackage{diagbox} 
\usepackage{ulem}

\newcommand{\makesupplementarytitle}{
    \begin{center}
        {\LARGE \textbf{Chimera: A Block-Based Neural Architecture Search \\[0.3em] Framework for Event-Based Object Detection}} \\[1em]
        {\large \textbf{Supplementary Material}} \\[1em]
    \end{center}
    \vspace{1.5em}
}

\graphicspath{ {./images/} }

\title{Chimera: A Block-Based Neural Architecture Search Framework for Event-Based Object Detection}

\author{Diego A. Silva\\
KAUST\\
 Thuwal, Saudi Arabia\\
{\tt\small diego.silva@kaust.edu.sa}
\and
Ahmed Elsheikh\\
Cairo University\\
Giza, Egypt\\
{\tt\small ahmed.elsheikh@eng.cu.edu.eg}
\and
Kamilya Smagulova\\
KAUST\\
 Thuwal, Saudi Arabia\\
{\tt\small kamilya.smagulova@kaust.edu.sa}
\and
Mohammed E. Fouda\\
Compumacy for AI Solutions \\
Cairo, Egypt\\
{\tt\small foudam@uci.edu}
\and
Ahmed M. Eltawil\\
KAUST\\
 Thuwal, Saudi Arabia\\
{\tt\small ahmed.eltawil@kaust.edu.sa}
}

\begin{document}
\maketitle
\begin{abstract}

Event-based cameras are sensors that simulate the human eye, offering advantages such as high-speed robustness and low power consumption. Established Deep Learning techniques have shown effectiveness in processing event data. Chimera is a Block-Based Neural Architecture Search (NAS) framework specifically designed for Event-Based Object Detection, aiming to create a systematic approach for adapting RGB-domain processing methods to the event domain. The Chimera design space is constructed from various macroblocks, including Attention blocks, Convolutions, State Space Models, and MLP-mixer-based architectures, which provide a valuable trade-off between local and global processing capabilities, as well as varying levels of complexity. The results on the PErson Detection in Robotics (PEDRo) dataset demonstrated performance levels comparable to leading state-of-the-art models, alongside an average parameter reduction of 1.6 times.


\end{abstract}    
\section{Introduction}
\label{sec:intro}

Object detection is a critical task in computer vision that involves identifying objects and determining their locations within an image. This capability is essential for various real-world applications, including autonomous driving \cite{object_detection_autonomous_driving}, robotics \cite{object_detection_robotics}, and surveillance \cite{object_detection_surveillance}. Traditionally, these applications rely on data from RGB cameras, which provide a continuous stream of high-resolution images \cite{survey_obj_det_deep_learning}. Recently, event-based cameras were introduced a new sensing paradigm, inspired by the human eye's functioning \cite{original_dvs_paper}. Unlike traditional cameras, pixels in event-based sensors generate outputs independently only when changes occur in the scene, leading to a spatio-temporal stream of events in response to brightness variations. Event-based sensors offer several advantages over RGB cameras, such as microsecond-range latency, a High Dynamic Range (HDR) exceeding 120 dB, power consumption in the milliwatt range, and potential memory savings by discarding redundant information \cite{eventbasedsurvey}. 

Among the various techniques developed for object detection using RGB input, deep learning algorithms, particularly the You-Only Look-Once (YOLO) family and transformer-based detectors—have achieved significant success \cite{survey_obj_det_deep_learning}. Various YOLO versions were introduced, enhancing its speed and accuracy while maintaining minimal trainable parameters \cite{survey_yolo}. There is a notable correlation between the success of deep learning methods in RGB applications and their performance in the event-based domain, as seen with convolutional networks \cite{red_megapixel}, \cite{astmnet}, \cite{reyolov8}, and transformer-based networks \cite{rvt}, \cite{sast}, \cite{get}, \cite{event-ssm}. Many of these networks are designed monolithically, meaning they consist of repeated layers of the same blocks. 

Additionally, in conventional computer vision, integrating different architectural blocks into a single hybrid network has shown benefits. Specifically, using convolutions in the trunk layers and transformers in the later stages has proven effective in balancing local and global contextual processing and managing computational complexity across various feature sizes \cite{fastervit}, \cite{maxvit}, \cite{edgevit}, \cite{efficientformer}. Some studies also explore combinations such as convolutional layers with MLP-Mixers \cite{convmlp}, and State-Space Models with Transformers \cite{mambavision}. Typically, the choice of combinations is influenced by researchers’ prior knowledge and experience. However, this process can be automatized by adopting Neural Architecture Search (NAS) frameworks \cite{ren2021comprehensive}. NAS employs a wide range of search and evaluation strategies, including gradient-based search, evolutionary algorithms, and reinforcement learning. These strategies can be further categorized into multi-shot, one-shot, and training-free Zero-Shot (ZS) NAS approaches. The first two require training multiple networks or a hyper-network, {which poses challenges for large search space exploration in real-world resource-constrained environments. Therefore, the utilization of ZS-NAS is more accessible and feasible.  Instead of full training which leads to high computational costs and sustainability concerns \cite{patterson2021carbon},} ZN-NAS uses proxy metrics for evaluation of the candidates, offering improved scalability, speed, cost-efficiency and sustainability.

 
{ The concept of a hybrid neural architecture featuring heterogeneous blocks can also be applied to the event-based processing domain. ZS-NAS is particularly well-suited for this task, as it minimizes the time and resources required to explore the vast design possibilities. The primary contribution of this work is introducing a scalable, two-stage NAS framework named Chimera, which aims to identify heterogeneous architectures specifically for event-based applications. This framework efficiently explores a broad design space within a limited time frame and {computing resources}, enhancing the potential for innovative and effective solutions in this area. The evaluation of the framework was done for PEDRo dataset.
 
\par
To the best of our knowledge, Chimera is the first framework focusing on NAS for the Event-Based Object Detection task. The structure of this paper is organized as follows: Section \ref{sec:related} offers background information on state-of-the-art neural architectures for object detection and on Zero-Shot NAS, which serves as the foundation for Chimera-NAS. Section \ref{sec:method} details the architecture of the Chimera network and the Chimera-NAS algorithm, including the design space and search metrics. This section also outlines the benchmarks developed for evaluating the framework. Section \ref{sec:results} presents the results and discusses the findings. Finally, Section \ref{sec:conclusions} summarizes the conclusions drawn from this work.}

\section{Related Works}
\label{sec:related}

\subsection{Event-Based Object Detection}


Currently, there are various neural architectures available for vision tasks. Existing event-based object detectors can be divided into two primary categories based on their processing approach: sparse models and dense models. Sparse models process input event streams asynchronously and include techniques like Graph Neural Networks (GNN) \cite{aegnn}, \cite{sun2023event_gnn}, \cite{dagr} and Spiking Neural Networks (SNN) \cite{hybrid-snn-ann}, \cite{spikingDenseNet}, \cite{ems-yolo}, \cite{spiking-retinanet}, \cite{spiking-yolov4}, \cite{attention-rpnsnn}, \cite{sfod}, \cite{easnn}. In contrast, dense models convert event streams into an intermediate format suitable for neural networks that process image-like features. The most common and effective configurations for dense neural networks are built using convolutional layers \cite{red_megapixel}, \cite{astmnet}, \cite{aec}, \cite{aed}, \cite{reyolov8}, as well as self-attention blocks and their variants \cite{rvt}, \cite{sast}, \cite{get}, \cite{ergo12}, \cite{event-ssm}. Additionally, several architectures integrate Recurrent Neural Networks (RNNs) to enhance temporal modeling \cite{red_megapixel}, \cite{rvt}, \cite{astmnet}. Notably, State Space Models (SSM) \cite{event-ssm} and Hierarchical Memory Networks (HMNet) \cite{hmnet} are also implemented in this context. Although significant has been done on sparse models, there is still a gap in performance between them and the dense approaches, which motivates adopting the latter in this work.



\subsection{Hybrid Neural Networks}

Combining diverse blocks into a hybrid architecture and leveraging their complementary features can enhance performance while achieving a balanced trade-off between computational complexity and global/local modeling \cite{convmlp}, \cite{mamba_original}. For example, transformer-based models are recognized for their state-of-the-art accuracy in vision applications \cite{attention}, but their high computational complexity can make processing high-resolution images challenging. To mitigate this issue, it is common practice to employ convolutional layers in the initial stages for input downsampling, followed by transformer-based blocks as the resolution decreases \cite{hassani2021escaping}. This approach helps to maintain a balance between local and global feature modeling throughout the network \cite{edgevit}, \cite{maxvit}, \cite{fastervit}, \cite{efficientformer}.

Similarly, convolutional layers have been used with MLP-Mixers to accommodate arbitrary input resolutions while reducing computational complexity \cite{convmlp}. In EfficientVMamba, an integration of convolutional blocks with State Space Models (SSM) was implemented, but unlike previous approaches, the SSM blocks were positioned in the early stages of the network to maximize global feature capture, with convolutional layers placed in the later stages \cite{efficientvmamba}. Conversely, MambaVision \cite{mambavision} employs convolutional layers at higher resolution layers while incorporating a mixer block that alternates between Mamba \cite{mamba_original}, an SSM block, and self-attention \cite{attention}. Other methodologies explore modifications of convolutional blocks with self-attention \cite{botnet}, the reverse \cite{crossvit}, \cite{coat}, \cite{visformer}, \cite{fastervit}, and even the creation of novel blocks that combine both paradigms \cite{visformer}, \cite{maxvit}, \cite{edgevit}, \cite{cvt}.

\par

\subsection{Zero-Shot NAS}
\label{sec:zershotnas}

Neural Architectural Search (NAS) was developed to automate the process of finding the structure and design of neural networks considering the given constraints to improve performance 
In this work, the preference is given to the Zero-Shot NAS, which eliminates the need for training neural networks and, therefore, improves cost and time efficiency \cite{zenscore}. Moreover, it offers high scalability and can be optimized for specific metrics using zero-shot proxies. The proxies are developed based on theoretical and empirical analysis of deep neural networks,  incorporating factors such as topology, initialization, gradient propagation, etc. Understanding how they impact the overall performance enhances interpretability and predictions. 
\par 
The implementation of Zero-Shot NAS requires identifying a design space of candidates $\mathcal{F}$ and selection of proxy metrics. As a result, the framework evaluates the candidate architectures, ranks them according to the estimated proxy scores, and selects the top architectures. 


\section{Methodology}
\label{sec:method}

Event-based cameras are 2D sensors that detect brightness fluctuations at the pixel level. This phenomenon can be quantitatively represented as follows:

\begin{equation}
\small
    \Delta L(x_{k}, y_{k}, t_{k}) \ge p_{k}C
    \label{eq:brightness_change}
\end{equation}

In this equation, $\Delta L$ represents the logarithmic change in the current of a photoreceptor (brightness) at the pixel coordinates $(x_{k}, y_{k})$ at timestamp $t_{k}$. The term $p_{k} \in \{+1, -1\}$ denotes the polarity of the event, indicating that a brightness change that surpasses the threshold (C) in absolute value triggers a positive or negative event \cite{original_dvs_paper}. Each event is defined by the tuple $e_{k} = (x_{k}, y_{k}, p_{k}, t_{k})$. To facilitate the processing of event data, it is common to convert it into an intermediate format that Deep Learning algorithms can easily handle. 
\par 
Chimera has a modular design consisting of fixed and variable computing blocks, also comprising choices of such event conversion. Overall, its structure is inspired by the architecture of Recurrent YOLOv8 (ReYOLOv8) \cite{reyolov8} and consists of a recurrent backbone module for feature extraction from the input and the multi-scale feature fusion and detection heads adopted from the original YOLOv8 model \cite{yolov8}. The NAS in Chimera focuses on finding the best-performing combination of heterogeneous variable blocks for composing the architecture of a recurrent backbone. 

The framework includes multiple steps:
\begin{compactitem}
\item Creation of a design space and benchmarks;
\item Selection of proxy metrics and evaluation of the candidates from the search space; 
\item Evaluation of the resulting Chimera network on the dataset PEDRo.
\end{compactitem}





\subsection{Chimera Network Organization}


\par 



\begin{figure}[ht]
    \centering
    \includegraphics[scale=0.54]{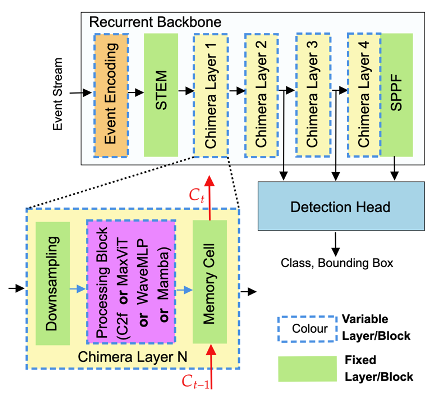}
    \caption{Structure of the Chimera Network.}
    \label{fig:chimera_overall}
\end{figure}
Figure \ref{fig:chimera_overall} displays the fundamental architecture of Chimera's recurrent backbone, which consists of seven layers. It begins with a variable Event Encoding block, followed by a 3x3 downsampling convolutional STEM layer. The subsequent four layers are called Chimera layers, each having an identical structure but varying compositions. Each of these four layers comprises three components: downsampling, processing, and a memory cell. The downsampling components resemble the STEM layer, while the memory cell is a fixed structure based on ConvLSTM \cite{convlstm}. The ConvLSTM is modeled after a standard LSTM \cite{gers2000learning}, with the critical distinction that the fully connected layers in the LSTM gates have been replaced by convolutional layers, allowing it to effectively process spatial features, similar to the implementations seen in Recurrent Vision Transformer (RVT) \cite{rvt} and ReYOLOv8 \cite{reyolov8}. The Downsample block diminishes the spatial dimensions before passing the features to the variable processing block, which extracts relevant information from the current features, while the following memory cell performs spatiotemporal modeling between the current and previous feature maps. The processing block can utilize any option available in Chimera's component library, and the choice of block for each Chimera Layer is made independently from the others.

The final layer of the recurrent backbone is a fixed Spatial Pyramid Pooling Fast (SPPF) \cite{spp} block, stacked to Chimera Layer 4 and is inherited from YOLOv8 \cite{yolov8}. Detection within the Chimera framework is carried out using the multi-scale YOLOv8 detection head, considering the findings reported in \cite{reyolov8} as well as the successful application of other YOLO models within the event-based literature \cite{rvt}, \cite{ergo12}, \cite{event-ssm}.

\subsection{Library of Components}
\label{sec:library}

The library supporting Chimera is comprised of various building blocks and options for data encodings. This section will provide a brief overview of each component. Further details regarding their implementations are available in the Supplementary Material.

\subsubsection{Building Blocks}
\label{sec:building_blocks}

\begin{itemize}
\item {\textbf{Convolutional Layers.}} The well-recognized capability of Convolutional Neural Networks (CNNs) to extract features has significantly transformed a variety of computer vision tasks \cite{krizhevsky2012imagenet}. For example, YOLOv8, which serves as the foundation of the Chimera framework, is composed of backbone, neck, and head blocks made entirely of convolutions, like the downsampling convolutions and the C2f blocks adopted for finer feature extraction \cite{yolov8}. Then, besides the convolutional layers on the YOLOv8 detection heads and in the downsampling and SPPF operations from the backbone, the C2f was selected as one of the possible processing blocks for the Chimera Layers.


\item {\textbf{Transformers.}} Transformers are highly powerful in modeling global context information due to the presence of self-attention operations \cite{attention}. However, this operation has quadratic complexity concerning the input size, which incurs computational burdens. In this context, Multi-axis Vision Transformer (MaxViT) \cite{maxvit} is a variation of self-attention with reduced computational complexity. Moreover, it was already successfully adopted in the event domain \cite{rvt}, which motivated us to include it in Chimera's library.

\item {\textbf{MLP-Mixers.}} Multilayer Perceptron (MLP)-Mixers model both local and global relationships through channel mixing and token mixing \cite{mlpmixer}. Token mixing captures spatial information, while channel mixing focuses on feature information. Particularly, WaveMLP is an MLP-Mixer that treats tokens as waves, incorporating amplitude and phase information and introducing a Phase-Aware Token Mixing module (PATM) \cite{wavemlp}. Due to its flexibility and reported performance, WaveMLP was included in the Chimera library.





\item {\textbf{State Space Models.}}
Grounded in continuous-time linear systems, these models have recently gained prominence for their efficiency in parallel processing. A variety of models adhering to this principle have emerged, mainly differing in their matrix representations. The Mamba block, which is included in the library, has attracted significant attention recently, both in the context of Large Language Models (LLMs) \cite{mamba_original} and in the vision domain \cite{mambavision}.
\end{itemize}


\subsubsection{Data Encodings}
\label{sec:encoding}

To allow a dense neural architecture to effectively process input events, these events must be converted into a grid-like format \cite{red_megapixel}. Various encoding schemes have been proposed, each varying in its capacity to capture events. In our study, we concentrate on the Volume of Ternary Event Images (VTEI) \cite{reyolov8}, Mixed-Density Event Stacks (MDES) \cite{mdes}, Stacked Histograms (SHIST) \cite{rvt}, and Temporal Active Focus (TAF) \cite{aed-taf} encodings. The VTEI format was chosen for its simplicity, high throughput, efficient compression, and positive results demonstrated in ReYOLOv8 \cite{reyolov8}. MDES shares similarities with VTEI and has shown promising performance in Depth Estimation \cite{mdes}. SHIST is included due to its utilization in multiple studies within the event-based object detection literature \cite{rvt}, \cite{event-ssm}, \cite{sast}. The mentioned formats are based on the accumulation of events across bins, while TAF employs a FIFO-based approach \cite{aed-taf}, thereby enhancing the diversity of the library.


\subsection{Chimera-NAS}
\label{sec:metrics:chimeraNAS}

\subsubsection{Search Metrics}
\label{sec:metrics}

\label{sec:metrics:zen}
 
Chimera-NAS is inspired by ZS-NAS, which provides the flexibility to choose proxies for selecting the best candidate neural blocks. In this study, we selected several ZS-NAS proxies from a range of existing options \cite{survey_zsnas}, including gradient-based accuracy proxies such as Zen-Score \cite{zenscore} and the Neural Tangent Kernel (NTK) Condition Number \cite{ntk}, along with naive proxies based on the number of parameters, denoted as \textit{\#Params}, and Multiply-Accumulate (MAC) operations. Additionally, a diversity index was introduced to ensure heterogeneity within the architecture. Together, these components address multiple objectives to achieve a balanced structure.



\begin{itemize}
\item {\textbf{Zen-Score}}
The expressive capacity of a network refers to its ability to effectively capture complex relationships within the input data.

For vanilla CNNs, it can be associated with Gaussian complexity according to 
\begin{equation}
\small
    \phi(f) = \log{E_{\mathbf{x}, \mathbf{\theta}} \|\mathbf{  \nabla_{\mathbf{x}}f(\mathbf{x}|\mathbf{\theta})}\|_{F}}
    \label{eq:phi-score}
\end{equation}
where $\mathbf{x}$ is the input, $\mathbf{\theta}$ the network parameters, and $f(.)$ is the network backbone with the last feature before the Global Average Pooling (GAP) operation. The formulation from Equation \ref{eq:phi-score} considers a network without Batch Normalization (BN) layers. However, this leads to problems such as overflow when applied to deep networks. The Zen-score solved this problem by introducing BN layers and considering their variance into the score computation \cite{zenscore}. Furthermore, to avoid adopting the backward propagation from Equation \ref{eq:phi-score}, they calculate the score according to the finite differential
\begin{equation}
\small
    \Delta = {E_{\mathbf{x}, \mathbf{\epsilon}} \|\mathbf{ f(\mathbf{x}) - f(\mathbf{x} + \mathbf{\alpha}\mathbf{\epsilon})}\|_{F}}
    \label{eq:zen-differential}
\end{equation}
where $\epsilon$ is a random disturbance and $\alpha$ is an adjust parameter for this noise. Then, the Zen-score is given by:
\begin{equation}
\small
    Zen(f) = \log(\Delta) + \sum_{i}\log(\sigma_{i})
    \label{eq:zen}
\end{equation}
where $i$ refers to the index of the BN layers, each with its respective standard deviation $\sigma_{i}$. Originally, both $\mathbf{x}$, $\mathbf{\theta}$, and $\mathbf{\epsilon}$ were taken from a standard Gaussian Distribution \cite{zenscore}. Also, in Chimera, $f(.)$ will consider the whole backbone block, including the SPPF block. 
\item{\textbf{Neural Tangent Kernel (NTK) Condition Number}}
The NTK (Neural Tangent Kernel) of a Neural Network can be expressed as follows:

\begin{equation}
\small
    \Theta(x,x') = \mathbf{J}(x)\mathbf{J}(x')^T
    \label{eq:ntk}
\end{equation}
where $\mathbf{J}(x)$ represents the Jacobian evaluated at (x) with respect to the parameters $\theta$. It has been demonstrated that the training dynamics of a neural network under gradient descent can be characterized by the conditioning of its NTK matrix, which is given by:

\begin{equation}
\small
    NTK_{cond} = \frac{\lambda_{0}}{\lambda_{m}}
    \label{eq:ntk_cond}
\end{equation}
where $\lambda_{0}$ and $\lambda_{m}$ denote the lowest and highest eigenvalues of $\Theta(x,x')$, respectively. A well-conditioned NTK matrix is linked to enhanced trainability, and a negative correlation between NTK conditioning and test-set accuracy for image classification tasks has also been reported \cite{ntk}.

\item{\textbf{Model Parameters and Complexity.}}
The number of parameters in a model and its complexity, measured in terms MACs, have been found to exhibit a certain degree of correlation with the model's test accuracy. These factors will also be included in the analysis of this work \cite{survey_zsnas}.

\item{\textbf{Diversity Index.}}
To assess the distribution of different types of blocks within a given architecture, we introduce an index to measure architectural diversity. Let $\textbf{B}(f) \in \mathbb{R}^{N \times 1}$ represent a vector where $N$ denotes the number of blocks available in a particular library. Each entry in this vector corresponds to the count of a specific block in the architecture $f$. The Diversity Index $D(f)$ for this architecture can be defined as follows:
\begin{equation}
\small
D(f) = 1 - \frac{\sum_{\forall{(i,j)}, j > i, i \neq j}{|\mathbf{B}(f)_{i} - \mathbf{B}(f)_{j}|}}{S}
    \label{eq:diversity_index}
\end{equation}
where ${\mathbf{B}(f)}_{i}$ represents the entries of the vector $\textbf{B}(f)$, and $S$ is a scale factor related to the maximum value that the summation term can attain. This formulation is designed such that when diversity is at its minimum—indicating that the architecture is homogeneous and consists of a single type of block—$D(f)$ is equal to $0$. Conversely, if there is no repetition in the block choices across all Chimera Layers, $D(f)$ will equal $1$.
\end{itemize}

\subsubsection{Search Space}

The design space $\mathcal{F}$ of Chimera includes a variety of modules and parameters, notably event encoding and processing blocks. The encoding blocks convert the input stream of events into formats such as VTEI, SHIST, MDES, or TAF, which are discussed in detail in Section \ref{sec:encoding}. For this study, the number of temporal bins assigned for the input representation is set to five. Following this, the output channel $Ch$ will be specified at the STEM layer, defined by a 3x3 convolution with a stride of 2.


\begin{table}[]
\caption{Chimera Design Space}
\footnotesize
\centering
\label{tab:chimera}
\begin{tabular}{|l|c|l|l|}
\hline
\multicolumn{1}{|c|}{\textbf{\#}} & \textbf{Layer}                                                                 & \multicolumn{1}{c|}{\textbf{Search focus}}                               & \multicolumn{1}{c|}{\textbf{Choices}}                                            \\ \hline
1                                 & \multicolumn{1}{l|}{Encoding}                                                  & Encoding type                                                           & \begin{tabular}[c]{@{}l@{}}VTEI, SHIST, \\ MDES, TAF\end{tabular}                \\ \hline
2                                 & \multicolumn{1}{l|}{STEM}                                                      & Output Channel $Ch$                                                                & 16, 24, 32, 40, 48                                                               \\ \hline
\multirow{5}{*}{3-6}              & \multirow{5}{*}{\begin{tabular}[c]{@{}c@{}}Chimera \\ Layers 1-4\end{tabular}} & Multiplier $M_i$                                                          & \begin{tabular}[c]{@{}l@{}}1, 1.25, 1.33, 1.50, \\ 1.66, 1.75, 2.00\end{tabular} \\ \cline{3-4} 
                                  &                                                                                & \begin{tabular}[c]{@{}l@{}}Block and \\ Block Params\end{tabular}     & \begin{tabular}[c]{@{}l@{}}C2f, MaxViT, \\ Mamba, \\ WaveMLP\end{tabular}   \\ \cline{3-4} 
                                  &                                                                                & \begin{tabular}[c]{@{}l@{}}Repeats \\ (Except Mamba)\end{tabular}  & 1, 2, 3                                                                          \\ \cline{3-4} 
                                  &                                                                                & \begin{tabular}[c]{@{}l@{}}Heads \\ (Mamba only)\end{tabular}           & 1, 2, 3                                                                          \\ \cline{3-4} 
                                  &                                                                                & \begin{tabular}[c]{@{}l@{}}Head multiplier \\ (Mamba only)\end{tabular} & 1.0, 1.25, 1.5, 2.0                                                              \\ \hline
\end{tabular}
\end{table}

For each Chimera layer, a multiplier $M_{i}$, a processing block, and its corresponding parameters $Block$ and $Block Params$ will be selected. The parameter $Repeats$ is applicable for all blocks, except Mamba. For the C2f, this parameter represents how many bottleneck blocks will be stacked inside its structure, and for the remaining ones, it represents how many of each structure will be stacked together. For Mamba, the number of heads will also be considered a parameter, and a head multiplier will scale the number of heads of a given Mamba block according to its predecessor. Considering all the possible choices, around 20,000 combinations can be generated from this model generator. Table \ref{tab:chimera} summarizes the Design Space for Chimera. More details about the implementation of each block can be find in the Supplementary Material.

\subsubsection{Chimera-NAS Algorithm}
\label{ref:chimera-nas}
The Chimera-NAS Algorithm operates in two stages. In the first stage, it generates and selects architectures using an Evolutionary Algorithm, favored for its simplicity and effectiveness, with prior results in the ZS-NAS domain \cite{zenscore}. The Fitness Score can be calculated based on any of the metrics described in Section \ref{sec:metrics}, or as a combination of these metrics. For a given design space $\mathcal{F}$, the optimization problem that Chimera-NAS seeks to solve to identify an effective architecture can be expressed as follows:

\begin{equation} \label{eq:chimera_optm}
\small
\begin{aligned}
& \underset{f \in \mathcal{F}}{\text{max}}
& & F = \alpha \mathbf{W}\cdot\mathbf{Z}(f) + (1- \alpha)D(f) \\
& \text{s.t.}
& & {Params(f)} \leq \text{$MAX_{\text{Params}}$}.
\end{aligned}
\end{equation}
where $\mathbf{Z}(f)$ $=$ $[Zen(f), MACs(f), NTK_{cond}(f)]$ is a vector that includes the ZS-NAS proxies for the architecture $f$, weighted according to the vector $\mathbf{W} \in \mathbb{R}^{3 \times 1}$; $D(f)$ represents the Diversity Index from Equation \ref{eq:diversity_index}; $Params(f)$ indicates the number of parameters in the architecture $f$; $MAX_{\text{Params}}$ is the upper limit on the number of parameters; and $\mathbf{\alpha} \in [0,1]$ is a constant that regulates the diversity of the architectures. {Kendall's Tau and Spearman's R are used to evaluate correlation between NAS proxies and ground truth. }

In the second stage, the top five architectures identified by the ZS-NAS are trained for 100 epochs. Ultimately, the architecture that achieves the highest mean Average Precision (mAP) is selected. This second step is crucial for reducing the inherent inaccuracies associated with proxy-based methods and fine-tuning the resulting architecture.

\begin{algorithm}
\footnotesize
    \caption{Chimera-NAS Algorithm}
    \begin{algorithmic}[1]
                           \STATE  \textbf{------------------------- First Stage: ZS-NAS  ----------------------------- }
        \STATE \textbf{Input:} Population size $N$, Search space $S$, Number of iterations $I$, Maximum Number of Parameters per architecture $Max_{Params}$, Population $P$, Fitness score $F$
        \STATE \textbf{Initialize:} $P \leftarrow \emptyset$
        
        \FOR{$i = 1$ \textbf{to} $N$}
            \STATE Create individual $I_i$
            \STATE Profile $I_i$ in terms of the ZS-NAS proxies
            \STATE Append $I_i$ to the population $P$
        \ENDFOR

        \FOR{$j = 1$ \textbf{to} $I$}
            \STATE Select a random individual $I_k$ from $P$, where $k = 1, \ldots, N$
            \STATE Apply Mutation to $I_k$, creating the individual $I_{k+1}$
            \IF{$Params(I_k) > Max_{Params}$}
                \STATE \textbf{goto} line 9 
            \ELSE
                \STATE Append $I_{k+1}$ to $P$
                \STATE Calculate the fitness score $F$ for $P$
                \STATE Remove the individual with the lowest $F$ from $P$
            \ENDIF
        \ENDFOR
        \STATE \textbf{\textbf---------------------- Second Stage: Fine-Tuning  ------------------------ }
        \STATE Choose the five individuals $I_k$ with the highest fitness scores and perform training for 100 epochs each
        \STATE \textbf{Output:} The individual $I$ with the highest mAP as the architecture output

    \end{algorithmic}
           \label{alghtm:chimera}

\end{algorithm}

\subsection{Dataset}
\label{sec:datasets}

 The PErson Detection in Robotics (PEDRo) dataset, focusing on robotics applications, is the dataset evaluated in this work. Captured in Italy using a handheld camera, PEDRo includes recordings of individuals in various scenes, lighting conditions, and weather scenarios, with 43k labels. The data was obtained using a DAVIS346 camera with a resolution of 346x260 pixels. To date, PEDRo is the only large-scale, real-world event-based dataset specifically tailored for robotics applications \cite{pedro}. In all instances, the events were aggregated into intermediate representations utilizing constant time windows of 40 ms, as adopted in previous works \cite{pedro}, \cite{reyolov8}. Results for another dataset, the Prophesee's Generation 1 Automotive Dataset (GEN1) \cite{gen1}, are included in the Supplementary Material.


\subsection{Chimera Benchmark}
\label{sec:benchmarking}

To evaluate the accuracy of the selected proxies in predicting test set performance, a benchmark was created using a subset of architectures from the Design Space. The PEDRo dataset was chosen for this evaluation due to its relatively small size. Each model was trained for 50 epochs, providing an effective compromise between runtime and convergence, and the mAP for the test set was recorded. Additionally, the Zen Score, MACs, number of parameters, and $NTK\_{cond}$ were documented. This analysis involved running 250 randomly generated heterogeneous models. Initially, each model was trained using VTEI, MDES, TAF, and SHIST representations, with 5 temporal bins each. 

Two correlation measures assessed how effectively the proxies approximate the mAP ground truth. One of these, Kendall's Tau, compares the ranking of models based on mAP and the proxies. The other metric used is Spearman's correlation, which measures the degree of monotonicity between the two variables, in this case, the proxies and the mAP \cite{survey_zsnas}.

\subsection{Training Procedure}
\label{methods:training}

For both the benchmark and final performance analysis, the same set of hyperparameters will be applied to each dataset, consistent with the procedures outlined in ReYOLOv8 \cite{reyolov8}. An image size of 320x256 was utilized for both datasets to enhance compatibility during the search process across different blocks and datasets. The models were trained using Stochastic Gradient Descent (SGD) with weight decay and a linear learning rate schedule. Given the recurrent nature of the architectures \cite{reyolov8}, Truncated Backpropagation Through Time (T-BPTT) \cite{tbptt} was employed. The benchmarks were trained for 50 epochs, while the final architectures were trained for 100 epochs after the searches were completed. All runs involving PEDRo, along with the executions of the Chimera-NAS algorithm, were conducted on a V100 GPU. Additional details can be found in the Supplementary Material.

\section{Results}
\label{sec:results}

\subsection{Evaluation of the Chimera Benchmark}
\label{sec:results:bench}

\subsubsection{Analysis of ZS-NAS proxies and event encodings}
\label{sec:proxies_and_encodings}
\begin{figure}[!t]
\centering
\begin{subfigure}{}
    \includegraphics[scale=0.5]{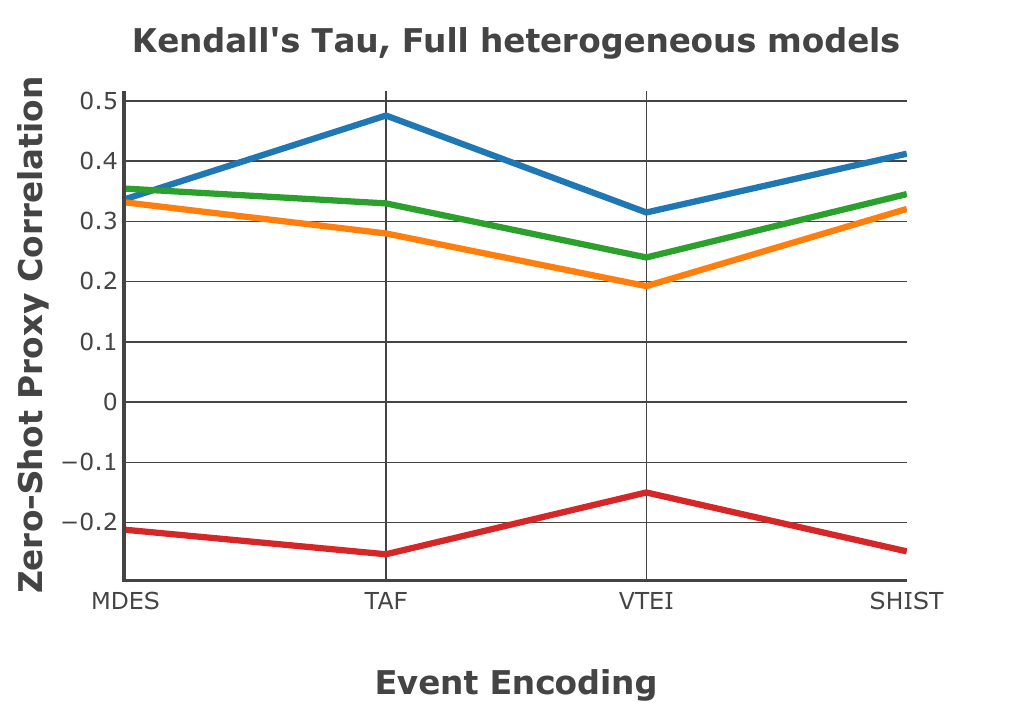}
\end{subfigure}
    \begin{subfigure}{}
        \includegraphics[scale=0.3]{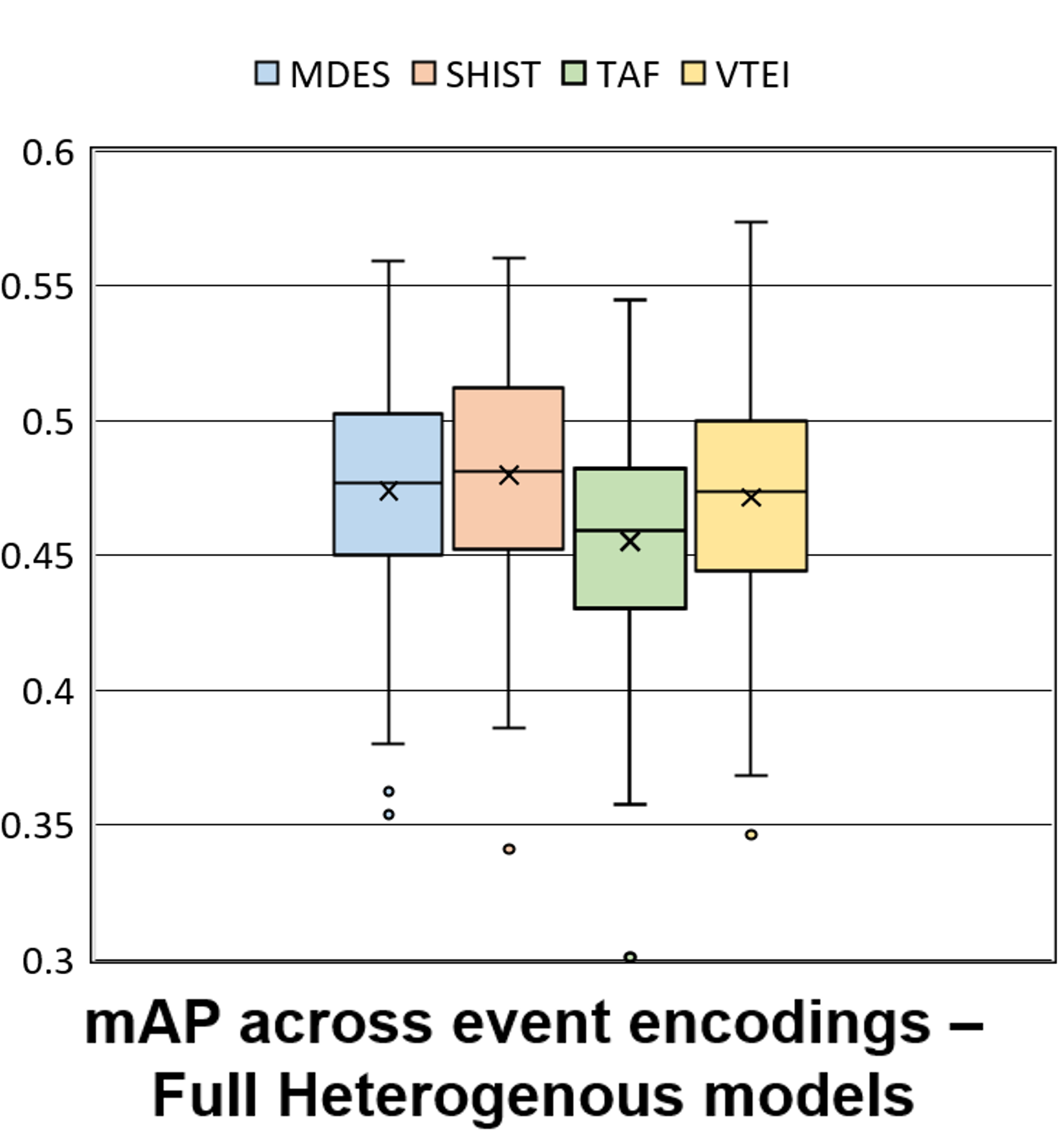}
    \end{subfigure}
\caption{a) Kendall's Correlation between the different proxies and data formats, for the full heterogeneous architectures.; b) Distribution of the mAPs for the different data encodings analyzed in the benchmark.}
\label{fig:cor_data_format}
\end{figure}

For the created benchmarks, the relationship between the ZS-NAS proxies and the event encodings was studied via analyzing the correlation between the rankings generated from these scores using Kendall’s correlation and their actual mAP rankings. The results are illustrated in Figure \ref{fig:cor_data_format}a. On average, the Zen Score demonstrated the best performance among other proxies, followed by the number of MACs and the total number of parameters. The NTK exhibited a negative correlation with the actual mAP, not exceeding 0.2 in absolute value. Furthermore, the computation time for NTK was significantly higher than that required for the Zen Score. Consequently, NTK was excluded from further analysis.




Figure \ref{fig:cor_data_format}b presents box plots of the mAP values for all heterogeneous architectures in the benchmark, grouped by encoding format. The results indicate that SHIST, MDES, and VTEI outperform TAF, whereas in the previous analysis, SHIST and TAF demonstrated a better relationship with the proxies than MDES and VTEI. Since SHIST demonstrated a superior performance in both evaluations, it was decided to use it for the subsequent experiments. 


Based on the results presented in Figure \ref{fig:cor_data_format}, a decision was made to explore further a fitness function that combines only the Zen Score and the number of MACs proxies. The number of parameters was excluded from the main function and instead treated as a constraint during the search process. {Table \ref{tab:max_correlation} reflects the impact of various weight combinations for the Zen Score and MACs on their corresponding Kendall's and Spearman's correlations relative to the mAP at 50 epochs, derived from the full heterogeneous subset of SHIST in the Chimera benchmark} 
The data reveals that the correlations increase up to a point when 60\% of the score is derived from the Zen Score and 40\% from MACs. Beyond this combination, the correlations begin to decline. Consequently, these weights were selected for use in the following experiments.

\begin{table}[]
\centering
\footnotesize
\caption{Kendall's and Spearmans' Correlations for different weighings of Zen Score and MACs for the SHIST subset of the full heterogeneous benchmark.}
\label{tab:max_correlation}
\begin{tabular}{|c|c|c|c|}
\hline
\begin{tabular}[c]{@{}c@{}}Zen\\ Weight\end{tabular} & \begin{tabular}[c]{@{}c@{}}MACs\\ Weight\end{tabular} & \begin{tabular}[c]{@{}c@{}}Kendall's\\ Tau\end{tabular} & \begin{tabular}[c]{@{}c@{}}Spearman's\\ R\end{tabular} \\ \hline
0                                                    & 1.0                                                   & 0.35                                                    & 0.50                                                   \\ \hline
0.1                                                  & 0.9                                                   & 0.39                                                    & 0.55                                                   \\ \hline
0.2                                                  & 0.8                                                   & 0.42                                                    & 0.60                                                   \\ \hline
0.3                                                  & 0.7                                                   & 0.45                                                    & 0.62                                                   \\ \hline
0.4                                                  & 0.6                                                   & 0.47                                                    & 0.65                                                   \\ \hline
0.5                                                  & 0.5                                                   & 0.48                                                    & 0.67                                                   \\ \hline
\textbf{0.6}                                         & \textbf{0.4}                                          & \textbf{0.49}                                           & \textbf{0.67}                                          \\ \hline
0.7                                                  & 0.3                                                   & 0.48                                                    & 0.66                                                   \\ \hline
0.8                                                  & 0.2                                                   & 0.46                                                    & 0.64                                                   \\ \hline
0.9                                                  & 0.1                                                   & 0.44                                                    & 0.62                                                   \\ \hline
1.0                                                  & 0.0                                                   & 0.41                                                    & 0.59                                                   \\ \hline
\end{tabular}
\end{table}

\subsubsection{Impact of the Diversity Index}
\label{sec:res:search}

Based on the findings from Section \ref{sec:results:bench}, SHIST was approved as the reference event encoding for the Chimera-NAS algorithm. Furthermore, as shown in Figure \ref{fig:cor_data_format}a and Table \ref{tab:max_correlation}, the weights were set to $\mathbf{W} = [0.6, 0.4, 0.0]$, which correspond to weights of Zen-Score, MACs and NTK proxies. To assess the distribution of different blocks across architectures and to determine whether diversity contributes to performance enhancements, multiple searches were conducted by varying the parameter $\alpha$ from Equation \ref{eq:diversity_index} within the range of 0.05 to 1.0. These searches were conducted using Algorithm \ref{alghtm:chimera} with 50 individuals, 1,000 iterations, and a maximum parameter limit $MAX_{\text{Params}}$ of 10M per architecture. To evaluate the effectiveness of the searches, the top 5 architectures for each $\alpha$ value were trained for 100 epochs, and their final mAP for PEDRo was averaged. 
According to the results, the best mAP performance was achieved at $\alpha = 0.05$, surpassing the other configurations by approximately +2.0. More details are presented in the Supplementary Material.







\subsection{Search Results}
\label{sec:results:search_archs}

To conduct the search of the architectures, the Algorithm \ref{alghtm:chimera} was repeated with the same population and iteration parameters, utilizing the $\textbf{W}$ obtained from Section \ref{sec:results:bench} and the $\mathbf{\alpha} = 0.05$ established in Section \ref{sec:res:search} for three constraints of $MAX_{\text{Params}}$ set to 3M, 5M, and 10M. Generally, after the search, the composition of the population regarding block types remained nearly identical across all individuals, with only minor variations in other parameters such as the number of channels, heads, and multipliers. Table \ref{tab:chimera_compostion} presents the blocks within the Chimera Layers obtained as a result of search for these three cases. After 1,000 iterations for each scenario, the population exhibited the same composition and arrangement of blocks, with differences confined to the number of channels and other parameters specific to the blocks.
\par 

To optimize the number of trainable parameters, Chimera-NAS recommends positioning the MaxViT block within the earlier layers of Chimera to enhance the capture of global data context. The C2f block is a convolutional component that accelerates processing and enhances model accuracy, primarily concentrating on local feature extraction. In contrast, the SSM module in the Mamba architecture is specifically designed to address both long- and short-term dependencies. Consequently, it can be inferred that the Chimera-3M model places a greater emphasis on spatio-temporal modeling compared to the Chimera-5M. Furthermore, the WaveMLP block in Chimera Layer 4 facilitates the encoding of both local and global features.

For the Chimera-10M model, the initial phase focuses on learning short- and long-term dependencies while engaging in local feature extraction. Global feature extraction occurs in Chimera Layers 3 and 4 through the WaveMLP and MaxViT blocks.

\begin{table}[]
\centering
\footnotesize
\caption{Composition of the Chimera Layer for architectures found with the Chimera-NAS algorithm under different parameters constraints.}
\label{tab:chimera_compostion}
\begin{tabular}{|c|c|c|c|}
\hline
\diagbox[width=11em]{Chimera Layer}{Params}& 3M      & 5M      & 10M     \\ \hline \hline
Chimera Layer 1& MaxViT  & MaxViT  & Mamba   \\ \hline
Chimera Layer 2 & Mamba   & C2f     & C2f     \\ \hline
Chimera Layer 3 & C2f     & Mamba   & WaveMLP \\ \hline
Chimera Layer 4 & WaveMLP & WaveMLP & MaxViT  \\ \hline
\end{tabular}
\end{table}

\subsection{Comparison with the state-of-the-art}
\label{sec:results:sota}

\begin{figure}
    \centering
    \includegraphics[scale=0.3]{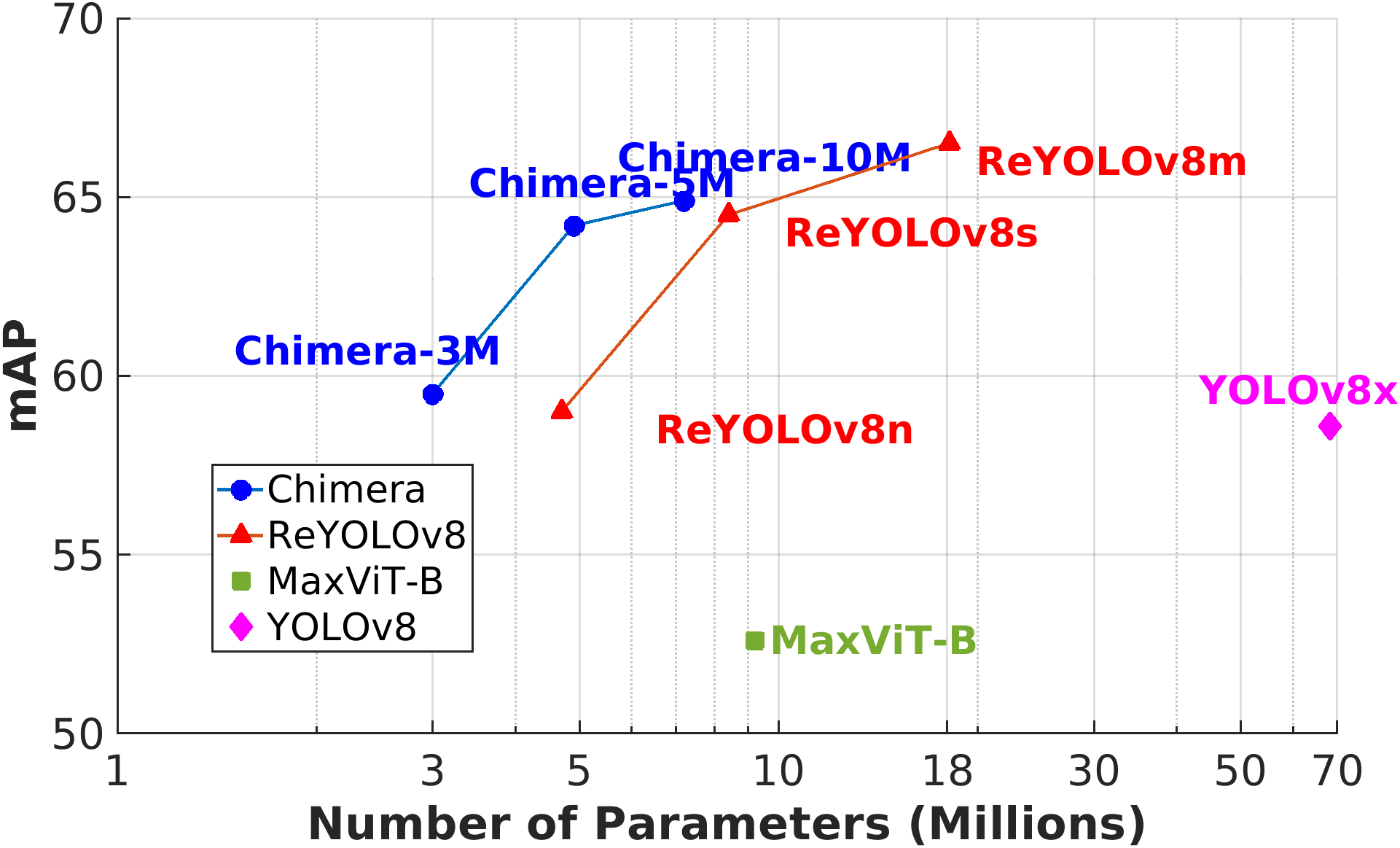}
    \caption{Comparison between the results and the state-of-the-art for the dataset PEDRo.}
    \label{fig:results}
\end{figure}

Figure \ref{fig:results} presents the architectures generated from the three searches—Chimera-3M, Chimera-5M, and Chimera-10M—alongside comparisons to other works in the literature for PEDRo. The results from ReYOLOv8 only include outcomes before the application of Random Polarity Suppression (RPS) \cite{reyolov8}, an augmentation technique specific to VTEI that has not yet been investigated alongside other formats, and therefore, were not applied to the Chimeras using SHIST. The MaxViT-Base model adheres to the backbone specifications from RVT \cite{rvt} but was trained under the same setup as the other models in this work.

The Chimera networks demonstrate strong performance, particularly at lower scales. Specifically, Chimera-3M performs similarly to ReYOLOv8n but with 1.5 times fewer parameters. Meanwhile, Chimera-5M shows an improvement in mAP of approximately 9\% compared to a similarly scaled ReYOLOv8n, achieving a comparable mAP to ReYOLOv8s, but with a model size that is 1.7 times smaller. However, Chimera-10M does not yield significant improvements over Chimera-5M or other similarly scaled models in the literature. One potential explanation for this is that at the 10M parameter constraint, the search space becomes larger than at lower scales, which may complicate the search process.

\section*{Conclusions and Future Work}
\label{sec:conclusions}

This work presents a two-stage NAS approach specifically aimed at Event-Based Object Detection. Instead of exploring variations of particular blocks, the architecture search centered on combining blocks from various paradigms within the literature to create more robust architectures. The resulting framework, named Chimera-NAS, utilizes proxies to assess architecture performance on test sets without requiring extensive training, enabling the examination of over 1,000 structures within a few hours. To manage the diversity of blocks within the architectures, a diversity index was introduced to quantify this aspect. The results obtained on the PEDRo dataset demonstrated significant improvements, achieving performance comparable to state-of-the-art models while maintaining an average parameter reduction of 1.6x. Future work will explore more blocks, alternative types of memory cells, such as State Space Models, and larger datasets, specifically Prophesee's GEN1 and 1MegaPixel.

\clearpage
\setcounter{page}{1}
\makesupplementarytitle

\section{Building Blocks Implementation}
\label{ref:supp:blocks}

 Section \ref{sec:building_blocks} of the manuscript discusses the motivation for utilizing C2f, MaxVit, WaveMLP, and Mamba blocks in building a hybrid architecture called Chimera. This Section of the Supplementary focuses on their structure and functionality.

\subsection{C2f Block}
\label{ref:supp:blocks:c2f}

\begin{figure}[ht!]
    \centering
    \includegraphics[width=0.45\linewidth]{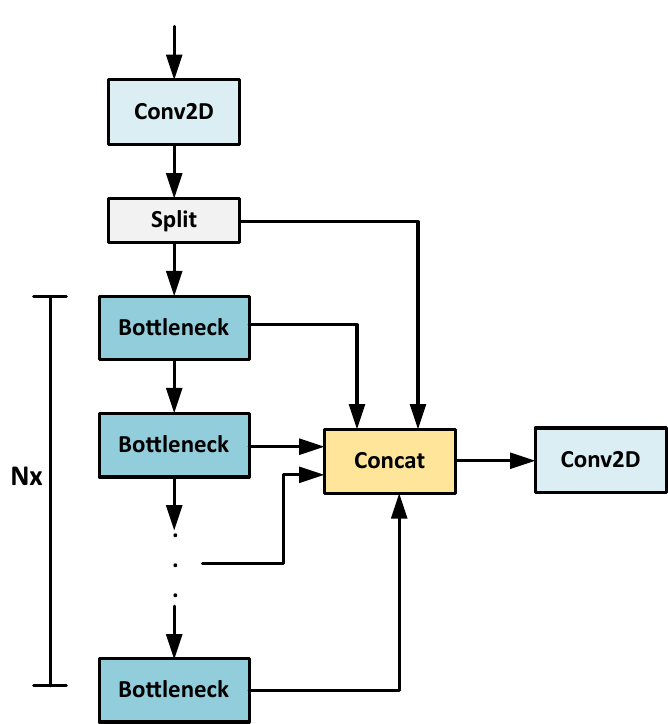}
    \caption{Structure of a C2f block.}
    \label{fig:c2f}
\end{figure}

Figure \ref{fig:c2f} depicts the architecture of the \textit{C2f} block \cite{yolov8}, an optimized variant of the Cross-Stage Partial (CSP) Bottleneck block featuring two convolutional layers. The first convolutional layer adapts the input channel count. Following this, a \textit{Split} block divides the features into two equal groups. One group is processed through a series of \textit{N} Bottleneck blocks, mirroring the structure of ResNet blocks. Importantly, the shortcut connections within these blocks are disabled in the PANET framework. The remaining split channels are combined with the outputs from each Bottleneck block, and a final convolution is applied to reduce the channel count. In this work, the parameters selected for the \textit{C2f} blocks are the number of input channels, output channels, and stacked bottlenecks.

\subsection{MaxViT Block}
\label{supp:blocks:maxvit}

\begin{figure}
    \centering
    \includegraphics[width=0.75\linewidth]{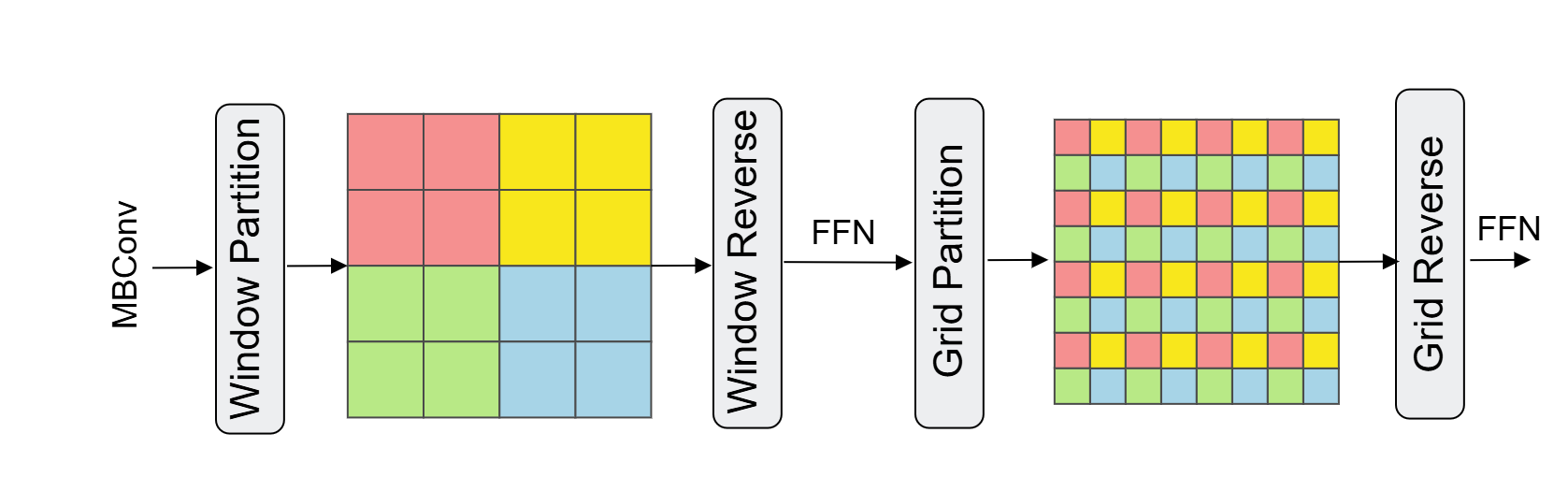}
    \caption{MaxViT block structure \cite{maxvit}.}
    \label{fig:maxvit_block}
\end{figure}

The Multi-Axis Vision Transformer (MaxViT) employs a multi-axis operation that separates the original self-attention into two distinct processes, thereby balancing global and local relationships while reducing computational complexity. The first process, Block Attention, segments the input into non-overlapping windows and applies self-attention within these windows, which captures local relationships. Subsequently, the grids are partitioned again to facilitate global modeling, performing self-attention in a dilated manner over the tokens \cite{maxvit}. Although the original study utilized an MBConv before the attention mechanism, this work adopts the approach from RVT, omitting the convolution \cite{rvt}. Figure \ref{fig:maxvit_block} illustrates this block. In this work, this block is instantiated only in terms of input and output channels. The remaining parameters are the same as those used in RVT \cite{rvt}, including the decision not to stack such blocks.


\subsection{Mamba Block}
\label{supp:blocks:mamba}

Mamba is a block based on State Space Models (SSM). A continuous-time SSM can be defined according to the following equations:

\begin{align}
 \quad & {h}^{'}(t) = \mathbf{A} {h}(t) + \mathbf{B}{x}(t)  \label{eq:ssm1}\\
 \quad & {y}(t) = \mathbf{C}{h}(t) 
\label{eq:ssm2}
\end{align}

In these equations, $h(t) \in \mathbb{R}^{M}$, $x(t) \in \mathbb{R}$, and $y(t) \in \mathbb{R}$ denote the state, input, and output vectors, respectively. The matrices $\mathbf{A} \in \mathbb{R}^{M \times M}$, $\mathbf{B} \in \mathbb{R}^{1 \times M}$, and $\mathbf{C} \in \mathbb{R}^{1 \times M}$ represent the state, input, and output matrices, respectively. Typically, these matrices are discretized by assuming a timescale $\Delta$ and applying a zero-order hold rule, as expressed by the following equations:

\begin{align}
 \quad & \mathbf{\bar{A}} = \exp(\Delta \mathbf{{A}})\label{eq:adist}\\
 \quad & \mathbf{\bar{B}} = (\Delta \mathbf{{A}})^{-1}(\exp{(\Delta \mathbf{{A}}}) - \mathbf{I})\cdot(\Delta \mathbf{{B}}) \label{eq:bdist}\\
 \quad & \mathbf{\bar{C}} = \mathbf{C} \label{eq:cdist}
\end{align}

In this way, the Equations \ref{eq:ssm1} and \ref{eq:ssm2} can be rewritten as:

\begin{align}
 \quad & {h}(t) = \mathbf{\bar{A}} {h}(t-1) + \mathbf{\bar{B}}{x}(t)  \label{eq:ssm1_d}\\
 \quad & {y}(t) = \mathbf{\bar{C}}{h}(t) 
\label{eq:ssm2_d}
\end{align}

Considering an input sequence with size $T$, a global convolution with kernel $\mathbf{{\bar{K}}}$ can be adopted to calculate Equations \ref{eq:ssm1_d} and \ref{eq:ssm2_d}, which can be replaced by

\begin{align}
 \quad & \mathbf{\bar{K}} = (\mathbf{C}\mathbf{\bar{B}}, \mathbf{C}\mathbf{\bar{A}}\mathbf{\bar{B}}, ..., \mathbf{C}{\mathbf{\bar{A}}}^{T-1}\mathbf{\bar{B}})
 \label{eq:ssm1_kernel}\\
 \quad & y = x*\mathbf{\bar{K}}
\label{eq:ssm2_kernel}
\end{align}

The efficiency of State Space Models (SSMs) is primarily influenced by the structure of the matrix $\mathbf{A}$, which is typically assumed to be diagonal. Mamba is an extension of the S4 SSM model that incorporates Selective Scan capabilities. To implement this, the parameters $\mathbf{B}$, $\mathbf{C}$, and $\Delta$ are adjusted to facilitate the filtering of inputs that are not relevant to the context, subsequently processing the pertinent inputs through a Scan operation \cite{mamba_original}.

\begin{figure}[ht!]
    \centering
    \includegraphics[width=0.3\linewidth]{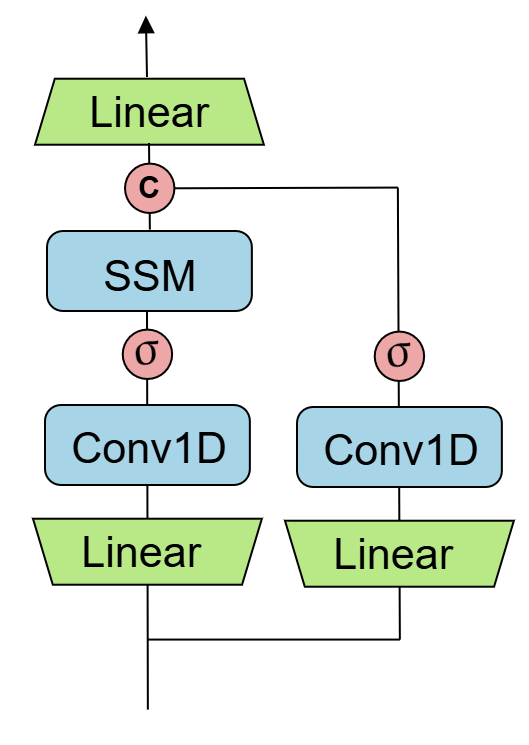}
    \caption{Mamba block. \cite{mambavision}}
    \label{fig:mamba_block}
\end{figure}

Figure \ref{fig:mamba_block} illustrates the implementation of the Mamba block for vision tasks. The architecture comprises two branches: one incorporating a State Space Model (SSM) and the other operating independently without it. In both branches, the input is initially projected through two linear layers, followed by a 1D convolution. After processing, the outputs from the two branches are concatenated and passed through an additional linear layer for final projection \cite{mambavision}. 

In MambaVision \cite{mambavision}, the convolutions are adapted to be single-directional, which is more suitable for vision applications. This modification enhances the ability of the model to capture relevant features without the constraints imposed by causality, improving its performance on visual data.

In the original implementation, the Mamba block alternates between a State Space Model (SSM) and Self-Attention mechanisms within the same stage. However, in this paper, we have decided to retain only the SSM block, as MaxViT already incorporates Self-Attention. This approach enables us to evaluate the effects of the Mamba block in a standalone manner within a specific stage. In this work, the Mamba block was instantiated based on the input and output channels, the window size used for reorganizing the inputs, and the number of stacked blocks.

\subsection{WaveMLP}
\label{ref:supp:blocks:wavemlp}

MLP-Mixers decompose the processing of tokens into two distinct tasks: Token Mixing and Channel Mixing. Token Mixing applies a multi-layer perceptron (MLP) across a specified channel for all tokens, enabling global interactions among them. In contrast, Channel Mixing, also known as Channel MLP, performs MLP operations across all channels for a given token, thereby enhancing local interactions \cite{mlpmixer}. 

\begin{figure}
    \centering
    \includegraphics[width=0.8\linewidth]{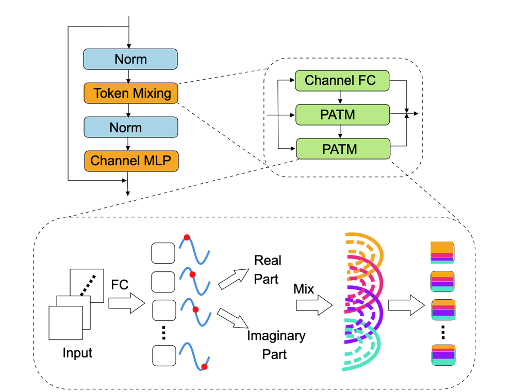}
    \caption{WaveMLP main block \cite{wavemlp}.}
    \label{fig:wavemlp_block}
\end{figure}

Figure \ref{fig:wavemlp_block} illustrates the structure of WaveMLP, a variant of the MLP-Mixer. This architecture interleaves Token Mixing and Channel MLP operations with Batch Normalization layers to ensure stable training. The Token Mixing component consists of three parallel branches: one Channel-MLP block and two Phase-Aware Token Mixing (PATM) blocks. In the PATM blocks, tokens are transformed into wave-like representations featuring real and imaginary components, with each PATM effectively mixing tokens along both the width and height axes \cite{wavemlp}. 

For this study, the parameters associated with WaveMLP included the input and output channels and the number of stacked blocks. Other parameters, such as patch size, stride, and padding, were fixed at $5$, $1$, and $2$, respectively.

\subsection{YOLOv8 Detection Head and PANET}
\label{supp:blocks:head}

\begin{figure}
    \centering
    \includegraphics[width=0.6\linewidth]{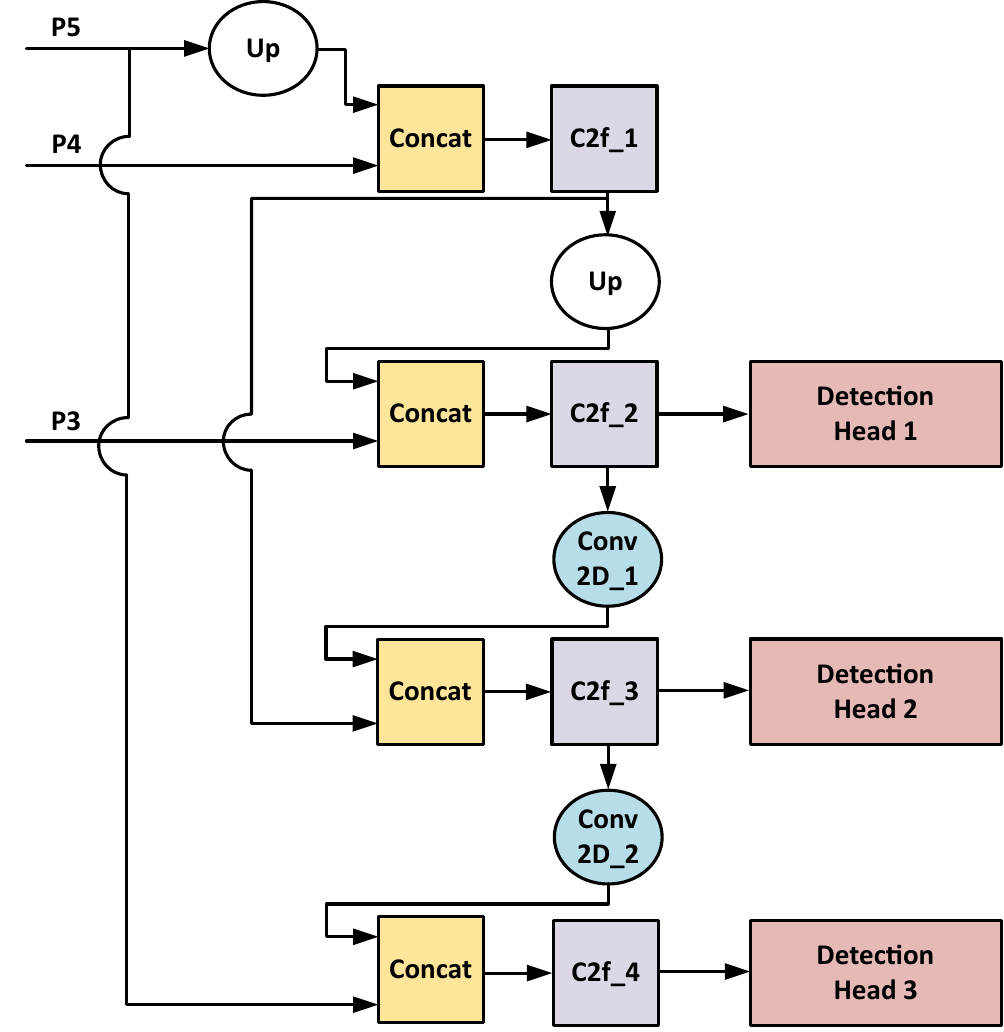}
    \caption{Multi-scale feature fusion and Detection Head structures from YOLOv8.}
    \label{fig:yolov8_panet}
\end{figure}

Like other algorithms in the YOLO family, YOLOv8 comprises structures responsible for feature extraction, multi-scale feature fusion, and a detection head. As illustrated in Figure \ref{fig:chimera_overall}, the last three feature maps from the backbone are forwarded to the Detection Head, which, for simplicity in this discussion, encompasses both the multi-scale feature fusion structure and the detection heads. The multi-scale feature fusion in YOLOv8 utilizes a Path Aggregation Network (PANET) that fuses those features from the backbone and transmits them to three detection heads. Figure \ref{fig:yolov8_panet} displays both structures, with $P5$, $P4$, and $P3$ denoting the last three feature maps from the backbone.

Figure \ref{fig:yolov8_panet} illustrates that the PANET includes four C2f blocks, two 2D convolutions, and three detection heads. The C2f blocks are derived from those depicted in Figure \ref{fig:c2f}, however implementing the Bottleneck blocks without residual connections. The $Conv_2D$ blocks represent $3\times 3$ downsampling convolutions, while the Detection Heads consist of various convolutional operations. Since this work primarily focuses on the design of backbones, the detection heads adhered to a templated design where the parameters for the C2f blocks, Conv2Ds, and Detection Heads were determined using a consistent set of rules applicable across all architectures. 

For all blocks, the input channels are defined based on the output channels of their corresponding preceding blocks. The output channels for the $C2f$ and $Conv_{2D}$ blocks must also be specified, as they influence the entire architecture. Specifically, the output channels of the $C2f$ blocks were set as multiples of the output channel of the STEM layer, $Ch_{0}$, following ratios similar to those used in YOLOv8n. Given the downsampling purpose of the $Conv_{2D}$ blocks, their output channels are the same as the input ones. The relationship between the output channels $Ch_{out}$ of each C2f block and $Ch_{0}$ is detailed in Table \ref{tab:panet_params}.

\begin{table}[ht!]
\centering
\caption{Output Channels adopted for the C2f blocks of the PANET.}
\label{tab:panet_params}
\begin{tabular}{|c|c|}
\hline
\textbf{Block} & \textbf{$Ch_{out}$} \\ \hline
C2f\_1         & 8$Ch_{0}$           \\ \hline
C2f\_2         & 4$Ch_{0}$           \\ \hline
C2f\_3         & 8$Ch_{0}$          \\ \hline
C2f\_4         & 1$6Ch_{0}$         \\ \hline
\end{tabular}
\end{table}

While the Detection Head and PANET are excluded from the search process, they still influence it. As illustrated in Algorithm \ref{alghtm:chimera}, Chimera-NAS imposes parameter limitations encompassing the entire model. Consequently, these components play a significant role in the model selection process.

\section{Event Encodings}
\label{supp:encodings}

Each event within an event stream arises from changes in the brightness and can be represented as a sequence  $e_k$ = ($x_k, y_k, t_k, p_k$) for $k$ = 1, 2, \ldots, $N$, where ($x, y$) denotes the pixel location, $t$ indicates the timestamp and $p$ reflects the polarity. A simple method for transforming an event stream into a dense, grid-like format involves stacking the events in various configurations. The event coding techniques implemented in this study are outlined in this section. In all the experiments, fixed time windows of 40 ms and 50 ms were adopted for PEDRo and GEN1 datasets, respectively.

\subsection{Mixed-Density Event Stacks}
\label{supp:encodings:mdes}

Due to the varying speeds of moving objects, stacking events according to a predefined number of events or time periods may result in information loss. For instance, short stacks may not effectively track slow-moving objects, while longer stacks with excessive events could overwrite fast-moving scenes. The Mixed-Density Event Stacks (MDES) format was proposed to address this issue, where the length of each event sequence $e_k$ is aggregated into $B$ stacks, each containing a different number of events per stack. The number of events within each stack $b \in [1, 2, \ldots, B]$ is given by $n = N/2^{b-1}$, where $N$ represents the total number of events available for the current time window. To enhance memory efficiency, for a given stack, only the last event on each pixel is considered, encoding positive, negative, and absent events with $255$, $127$, and $0$, respectively \cite{mdes}. In this work, we adopted MDES with $B = 5$.


\begin{figure}[hh]
    \centering
    \includegraphics[width=0.85\linewidth]{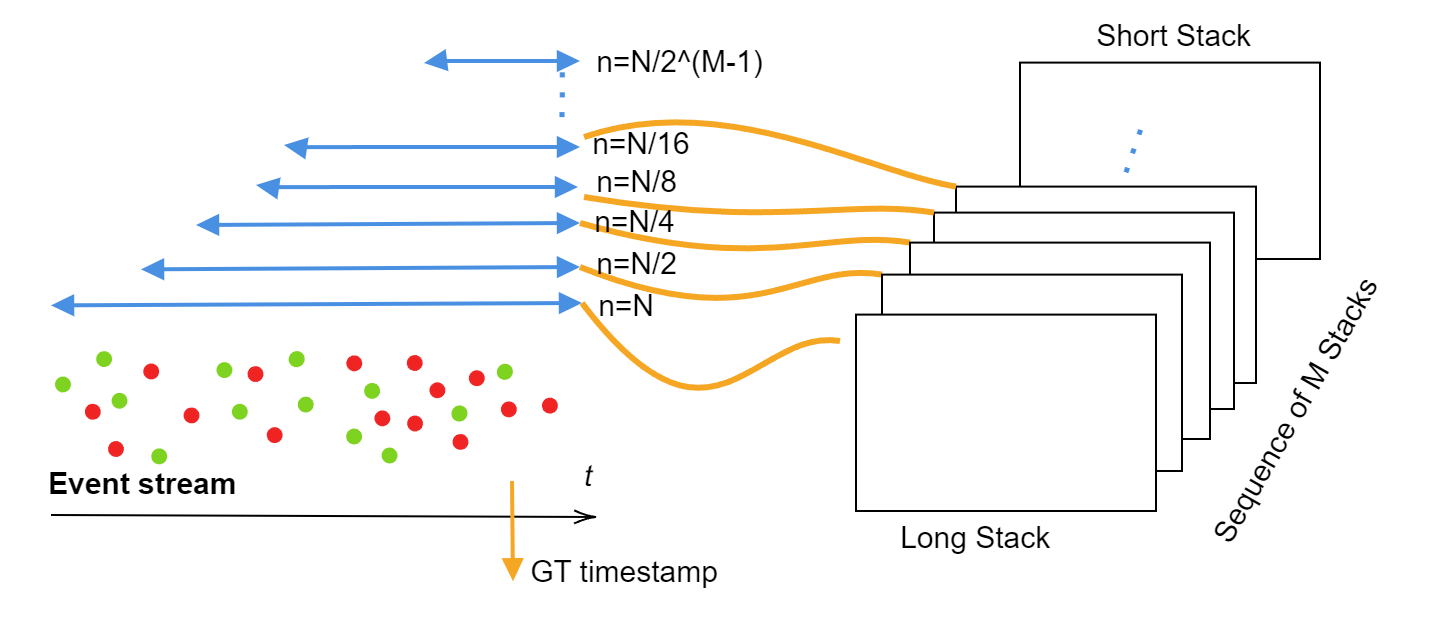}
    \caption{Mixed-Density Event Stacks (MDES).}
    \label{fig:mdes}
\end{figure}

\subsection{Stacked Histograms}
\label{supp:encodings:shist}

A Stacked Histogram (SHIST) is designed to save memory and bandwidth. SHIST is based on generating a 4-dimensional byte tensor from a given chunk of an event stream. The first two dimensions of this tensor correspond to the polarities and the $T$ discretization steps of time, while the last two dimensions represent the height $H$ and width $W$ of the camera. For a time window $[t_a, t_b)$, the SHIST tensor $\mathcal{E}$ can be represented as

\begin{equation}
E(p, \tau, x, y ) = \sum_{e_k \in \mathcal{E}}\delta(p-p_k)\delta(x-x_k,y-y_k)\delta(\tau-\tau_{k})
\end{equation}
where $\tau_k = \frac{(t_k - t_a) T}{(t_b - t_a)}$. To adapt it to the most adopted neural networks, the polarity and time dimensions are merged into a single one, changing the shape to $(2T, H, W)$ \cite{rvt}. In this work, we adopted SHIST with $T = 5$.

\subsection{Temporal Active Focus}
\label{supp:encodings:taf}
Typically, object detectors process output with a sampling period $\Delta \tau$. To avoid excessive data processing of all event spikes, the Temporal Active Focus (TAF) method concentrates only on the most recent non-zero $K$ events at each spatial and polar position \cite{aed-taf}. First-In-First-Out (FIFO) sliding queues of events $FIFO(p, t, x, y)$ with depth $K$ form a compact and dense tensor $S \in \mathbb{R}^{2K \times H \times W}$ containing the most meaningful data. In this work, we adopted TAF with $K = 5$.


\subsection{Volume of Ternary Event Images}
\label{supp:encodings:vtei}
The Volume of Ternary Event Images (VTEI) method ensures high sparsity, low memory usage, low bandwidth, and low latency. Similar to MDES, VTEI emphasizes the encoding of the latest event data but utilizes uniform temporal bin sizes and considers event polarity, +1 and -1. The VTEI tensor is created in several steps. The first step involves initializing a tensor $I$ with dimensions $B \times H \times W$, where $B$ represents the temporal bins, while $H$ and $W$ are the height and width of the camera. Next, an event stream consisting of $N$ events is sampled over a consistent time window $[t_a, t_b)$:
\begin{equation}
T_k = \frac{(t_k-t_a)B}{(t_b-t_a)}
\end{equation}
where $t_a$ and  $t_b$ are the initial and final timestamps, and $T_k$ is the temporal bin assigned for the timestamp $t_k$. In this work, we utilized $B=5$.


\begin{figure}[hh]
    \centering
    \includegraphics[width=0.85\linewidth]{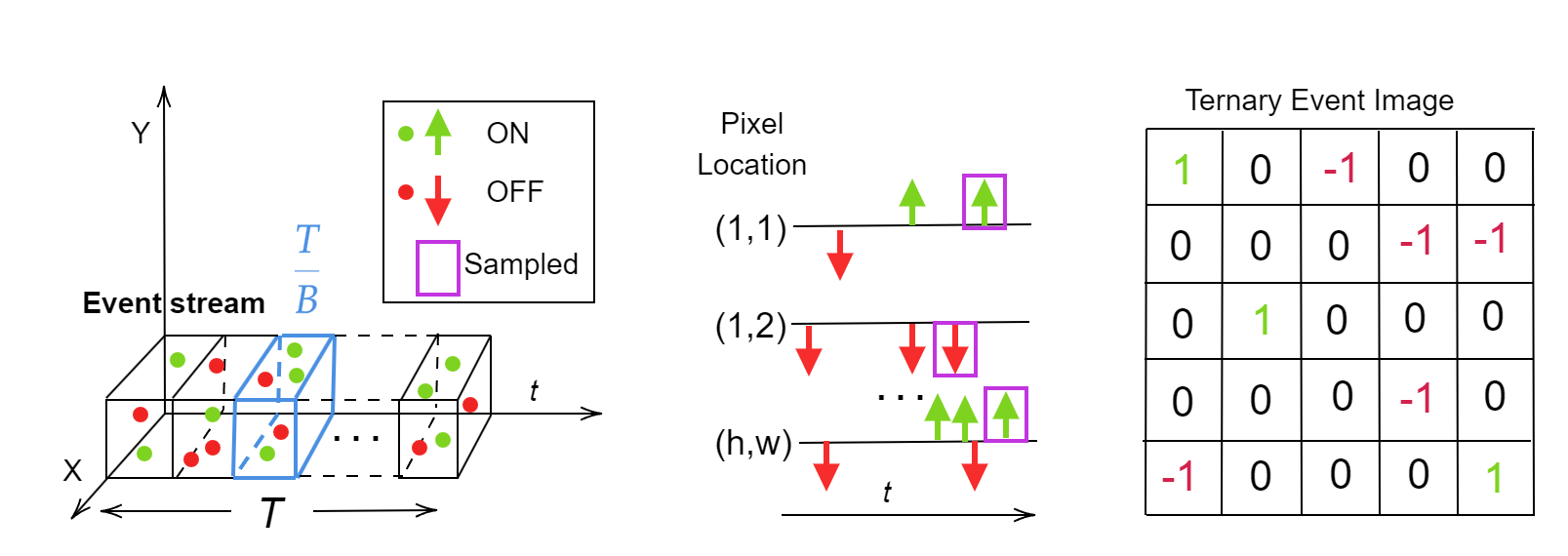}
    \caption{Volume of Ternary Event Image (VTEI).}
    \label{fig:VTEI}
\end{figure}

\section{More details about training}
\label{supp:training}

The training hyperparameters and procedures implemented in this study were primarily adapted from ReYOLOv8 \cite{reyolov8}, and YOLOv8 \cite{yolov8}, with minor modifications to batch sizes and learning rates. Table \ref{tab:hyps_training} summarizes the hyperparameters used for all runs on PEDRo and GEN1. $LR0$ denotes the initial learning rate, while $LRf$ signifies the final learning rate by a linear learning rate schedule. The models were optimized using Stochastic Gradient Descent (SGD) with a momentum of 0.937. Simple grid searches were adopted for the hyperparameters that differ from the literature.

Regarding data augmentation, \textit{HFLlip} refers to horizontal flipping, while \textit{Zoom-Out} was applied with ratios ranging from 1.2 to 1.0. A warmup period of 3 epochs was implemented, featuring a learning rate bias of 0.1 and a warmup momentum of 0.8. The loss functions maintained the same parameters from YOLOv8 \cite{yolov8} alongside the confidence thresholds and non-maximum suppression parameters.

Although features like Automatic Mixed Precision and Half Precision are available within the YOLOv8 framework, they were not utilized in this work due to compatibility issues with certain blocks.

\begin{table*}[ht!]
\centering
\caption{Training Hyperparameters GEN1 and PEDRo}
\label{tab:hyps_training}
\begin{tabular}{|c|c|c|c|c|c|c|c|c|}
\hline
\textbf{Dataset}       & \textbf{Model} & \textbf{\begin{tabular}[c]{@{}c@{}}Sequence\\ Length\end{tabular}} & \textbf{\begin{tabular}[c]{@{}c@{}}Batch\\ Size\end{tabular}} & \textbf{\begin{tabular}[c]{@{}c@{}}Weight \\ Decay\end{tabular}} & \textbf{LR0} & \textbf{LRf} & \textbf{HFlip} & \textbf{\begin{tabular}[c]{@{}c@{}}Zoom\\ Out\end{tabular}} \\ \hline
\multirow{3}{*}{GEN1}  & Chimera-3M     & 11                                                                 & 48                                                            & 0.011                                                            & 0.03         & 0.0003       & 0.5            & 0.5                                                         \\ \cline{2-9} 
                       & Chimera-5M     & 11                                                                 & 48                                                            & 0.011                                                            & 0.03         & 0.0003       & 0.5            & 0.5                                                         \\ \cline{2-9} 
                       & Chimera-10M    & 11                                                                 & 48                                                            & 0.011                                                            & 0.03         & 0.0003       & 0.5            & 0.5                                                         \\ \hline
\multirow{3}{*}{PEDRO} & Chimera-3M     & 5                                                                  & 36                                                            & 0.005                                                            & 0.07         & 0.0007       & 0.5            & 0.2                                                         \\ \cline{2-9} 
                       & Chimera-5M     & 5                                                                  & 36                                                            & 0.005                                                            & 0.07         & 0.0007       & 0.5            & 0.2                                                         \\ \cline{2-9} 
                       & Chimera-10M    & 5                                                                  & 36                                                            & 0.005                                                            & 0.07         & 0.0007       & 0.5            & 0.2                                                         \\ \hline
\end{tabular}
\end{table*}

\section{Analyses of the Benchmark}
\label{supp:design_space}


By examining the Design Space presented in Table \ref{tab:chimera}, disregarding the event encodings yields approximately 5,000 potential backbones. Based on that, we randomly generated models within this set, resulting in about 250 models, which represents 5\% of the initial Design Space. We took care to maintain a uniform distribution throughout the possible choices available. Furthermore, by combining the four distinct event encodings with each backbone, we analyzed approximately 1,000 complete heterogeneous combinations.

\subsection{Performance across different models}
\label{supp:deisng_space:performance_models}

\par
\begin{figure}[ht!]
    \centering
    \includegraphics[width=0.6\linewidth]{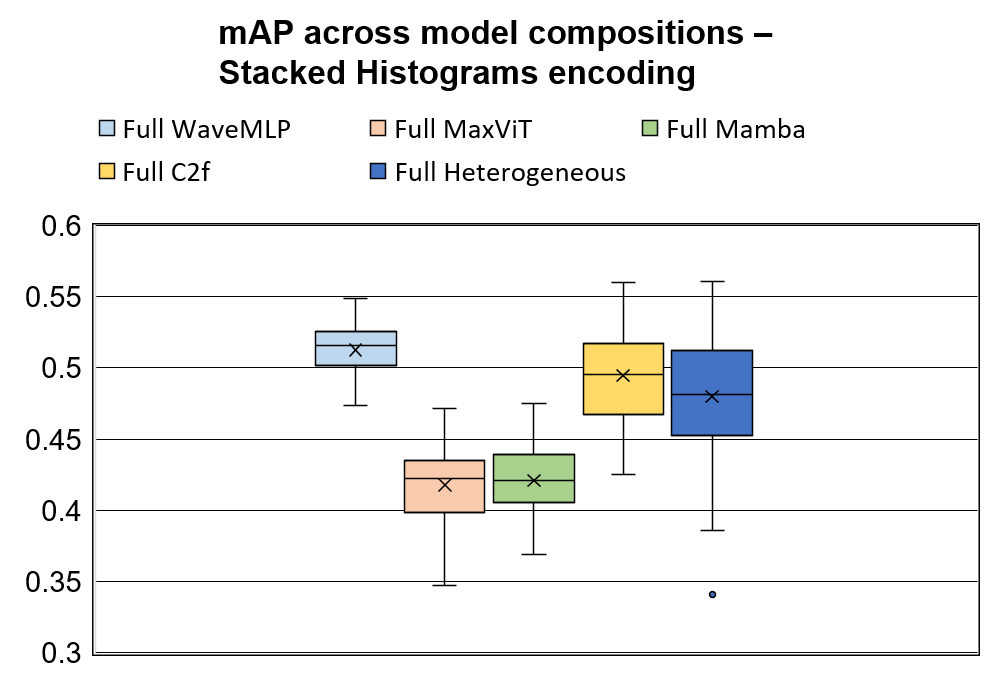}
    \caption{mAP distribution for the different model compositions.}
    \label{fig:map_compositions}
\end{figure}

 
In addition to the heterogeneous models used in the benchmark, we also randomly generated homogeneous models, which feature identical processing blocks across all Chimera Layers. These homogeneous models were intended to facilitate understanding of how individual components can influence the ZS-NAS proxies and overall performance. Approximately 50 models were created for each category. Like the fully heterogeneous models in the benchmark, these homogeneous models were trained for 50 epochs, and their mean Average Precision (mAP) was compared. Figure \ref{fig:map_compositions} displays the distributions across different categories, utilizing Stacked Histograms for encoding.

The results indicate that, on average, Full WaveMLP models perform better, followed by Full C2f, Full Heterogeneous, Full Mamba, and Full MaxViT. Notably, Full C2f and Full Heterogeneous models exhibited a more comprehensive range of performance and achieved higher mAP values than Full WaveMLP models when the maximum achieved values are compared.

\begin{figure}[ht!]
    \centering
    \includegraphics[width=0.6\linewidth]{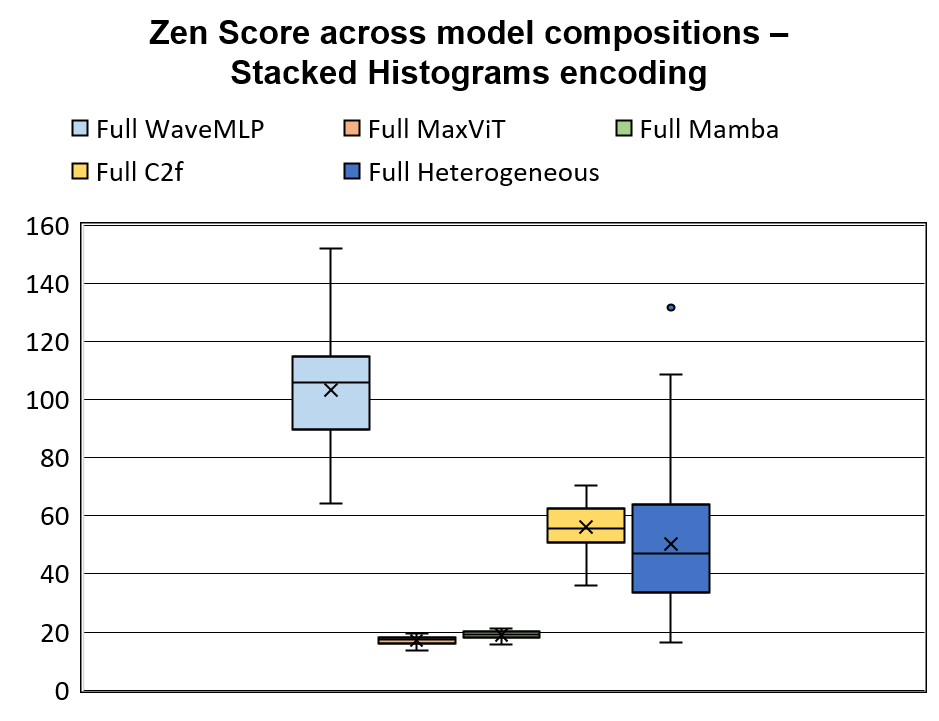}
    \caption{Distribution of the Zen Score for the different categories of models.}
    \label{fig:zen_models}
\end{figure}

Figure \ref{fig:zen_models} displays the distribution of the Zen Score across different categories. The average ranking among the categories mirrors the order shown for the mAPs in Figure \ref{fig:map_compositions}. However, the magnitude difference for Full WaveMLP models compared to other models is significantly greater than the proportional differences in mAPs. As discussed in Section \ref{sec:metrics:zen}, the Zen Score is derived from specific assumptions about the model type, which may explain why Full WaveMLPs receive higher scores while Full Mamba and Full MaxViT models do not. To address this bias towards Full WaveMLP blocks, the Diversity Index introduced in Section \ref{sec:metrics} will be examined further in Section \ref{supp:diversity}.

\subsection{Relationship between data format, model, and performance}
\label{supp:deisng_space:performance_formats}

The findings of the study examining the correlation between event encoding and Zen-Score are shown in Figure \ref{fig:cor_data_format} of the manuscript, highlighting significant variations in mAP at 50 epochs across different formats. Among these, SHIST exhibited the highest average performance, followed by MDES, VTEI, and TAF in that order. Additionally, the proxies exhibit varying behaviors depending on the format. Therefore, SHIST was chosen as the fixed data format. However, identifying metrics that facilitate efficient co-design of architectures and event encodings is essential. 

To support this claim, all models benchmarked—including the homogeneous models discussed in Section \ref{supp:deisng_space:performance_formats}—were compared across each data format considered in this work. Table \ref{tab:format_lead} presents the distribution of data formats based on the count of models where they showed the highest performance. The ordering observed in Figure \ref{fig:cor_data_format} is also reflected here. However, despite exhibiting better average performance and a stronger correlation with the ZS-NAS proxies, models utilizing SHIST achieved optimal results only 36\% of the time. Thus, without a metric that effectively captures the relationship between architectures and event encodings, numerous opportunities remain unexplored.

\begin{table}[ht!]
\centering
\caption{Event Encodings and the number of models on where they achieved the highest performance.}
\label{tab:format_lead}
\begin{tabular}{|c|c|}
\hline
\textbf{Event Encodings} & \textbf{Lead Performance} \\ \hline
SHIST           & 163                       \\ \hline
MDES            & 147                       \\ \hline
TAF             & 17                        \\ \hline
VTEI            & 126                       \\ \hline
\end{tabular}
\end{table}


\section{Diversity Index Analysis}
\label{supp:diversity}

Figure \ref{fig:zen_models} illustrates that the Zen Score is biased towards Full WaveMLP models, which might cause the search algorithm to favor these over heterogeneous models. To address this concern, we developed the Diversity Index from Equation \ref{eq:diversity_index} and incorporated it, along with a parameter $\alpha$, into the cost function specified in Equation \ref{eq:chimera_optm}. This adjustment was meant to ensure that the architectures maintain a certain degree of diversity. To assess the impact of the Diversity Index and $\alpha$ in the search, Chimera-NAS was executed with 11 different $\alpha$ values ranging from 0.1 to 1.0 in increments of 0.1, alongside a population of 50 individuals, 1,000 iterations, and a maximum parameter count of 5M. Additionally, a point at $\alpha = 0.05$ was included. The top five performing architectures for each alpha were then trained for 100 epochs. 

\begin{figure}
    \centering
    \includegraphics[width=0.65\linewidth]{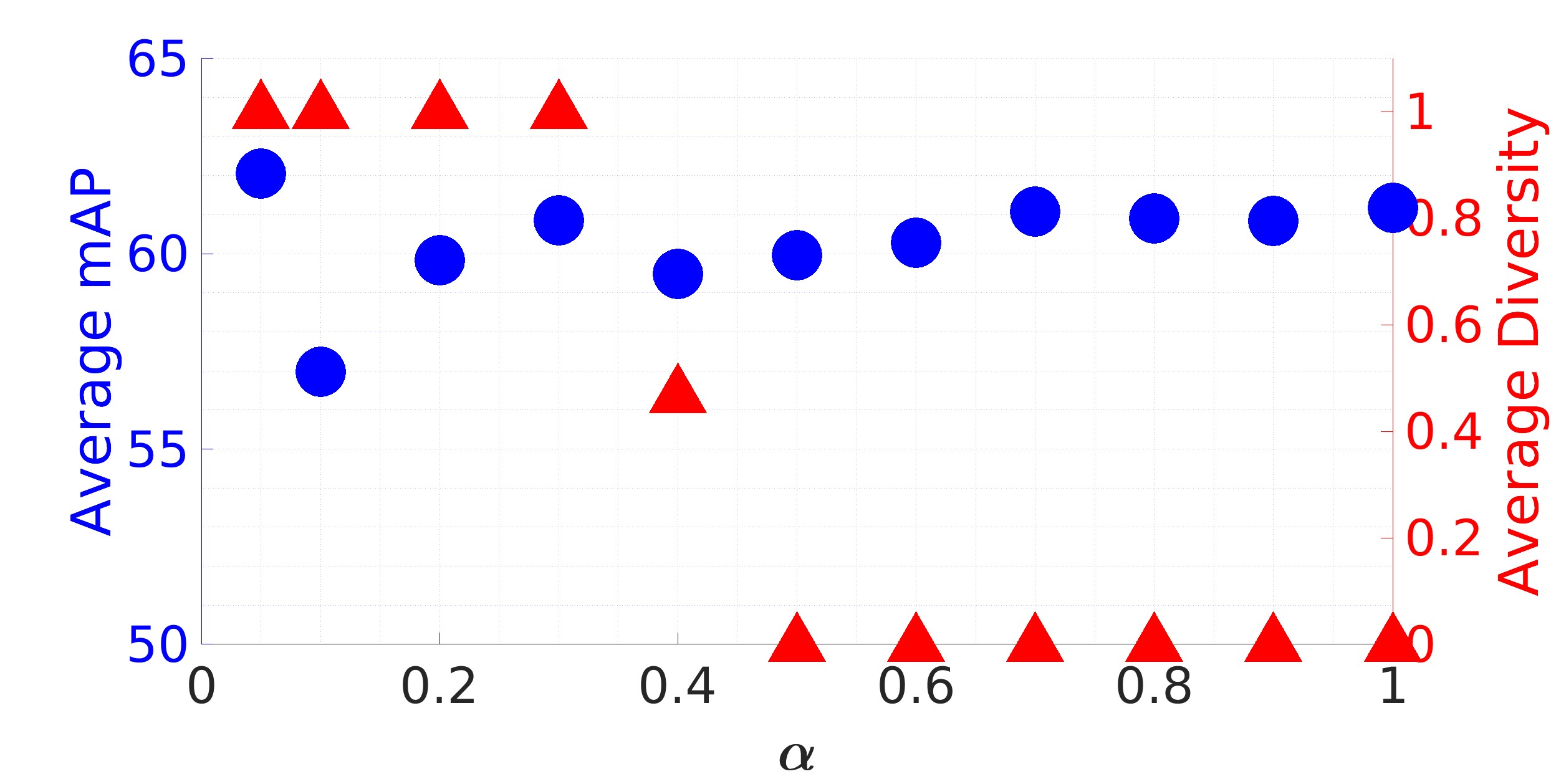}
    \caption{Average mAP of the Top-5 performers and Average Diversity of populations after evolving over $\alpha$ -5M}
    \label{fig:div_vs_alpha}
\end{figure}

Figure \ref{fig:div_vs_alpha} illustrates the average Diversity Index of the top-five architectures as a function of alpha. The figure indicates that the search process can be divided into two distinct regions: one from 0.05 to 0.4, where architectures exhibit high diversity, and another from 0.4 onwards, where diversity drops to zero, indicating that the blocks consist solely of a single block type. In these cases, all blocks were composed of WaveMLPs, as anticipated based on the Zen Score distribution shown in Figure \ref{fig:zen_models}.

When examining performance and the average mAP across the scenarios, the best results were observed at $\alpha = 0.05$. This suggests that, given the current constraints, optimal performance is achieved when diversity is maximized, with the ZS-NAS scores serving as a differentiation mechanism among architectures featuring similar block structures.



\section{Chimera-NAS Runtime}
\label{supp:algth_runtime}

Table \ref{tab:chimeranas} presents the runtime for the various steps involved in Chimera-NAS. This analysis considered a population of 50 individuals, 1,000 iterations, and a maximum parameter count of 5M per model. Stage 1 implements the ZS-NAS step from Algorithm \ref{alghtm:chimera}. This stage was conducted on an Ubuntu OS with 265GB of RAM, an Intel Xeon Gold @2.10GHz x 104 processor, and a Quadro RTX 4000 GPU. Under this setup, Chimera-NAS evaluated 1,050 models in 1.32 hours, which is 2.64 times faster than a single complete training session of 100 epochs on the PEDRo dataset using an NVIDIA V100 GPU, as also shown in the table. 

Since the combined score of MACs and Zen Score demonstrated a maximum Kendall correlation of 0.50, as indicated in Table \ref{tab:max_correlation}, an additional step was included—training the top five architectures for 100 epochs each. For PEDRo, this extra training step on an NVIDIA V100 GPU requires 17.4 hours.
\begin{table}[ht!]
\centering
\caption{Runtimes related to Chimera-NAS}
\label{tab:chimeranas}
\caption{Analysis of the runtimes involved on Chimera-NAS}

\begin{tabular}{|c|c|}
\hline
\textbf{Item}                                                                   & \textbf{Runtime} \\ \hline
\begin{tabular}[c]{@{}c@{}}Stage 1 \\ (ZS-NAS) \end{tabular}                & 1.32 hours       \\ \hline
\begin{tabular}[c]{@{}c@{}}Training - 100 epochs \\ (v100 GPU) \end{tabular} & 3.48 hours       \\ \hline
\begin{tabular}[c]{@{}c@{}}Stage 2 \\ (Train top-5 from ZS-NAS)\end{tabular}                      & 17.4 hours       \\ \hline
\end{tabular}
\end{table}

\section{Details about the architectures}
\label{supp:details}

Tables \ref{tab:chimera3m}, \ref{tab:chimera5m}, and \ref{tab:Chimera10M} shows more details regarding the Chimera models, such as the number of channels and repeats.

\begin{table}[ht!]
\centering
\caption{Chimera-3M Detailed Architecture}
\footnotesize
\label{tab:chimera3m}
\begin{tabular}{|c|c|c|c|c|}
\hline
\textbf{Block}                                             & \textbf{\begin{tabular}[c]{@{}c@{}}Processing\\ Unit\end{tabular}} & \textbf{\begin{tabular}[c]{@{}c@{}}Input \\ Channel\end{tabular}} & \textbf{\begin{tabular}[c]{@{}c@{}}Output\\  Channel\end{tabular}} & \textbf{Repeats} \\ \hline
STEM                                                       & Conv2D                                                             & 10                                                                & 16                                                                 & 1                \\ \hline
\begin{tabular}[c]{@{}c@{}}Chimera \\ Layer 1\end{tabular} & MaxViT                                                             & 16                                                                & 32                                                                 & 1                \\ \hline
\begin{tabular}[c]{@{}c@{}}Chimera\\ Layer 2\end{tabular}  & Mamba                                                              & 32                                                                & 48                                                                 & 1                \\ \hline
\begin{tabular}[c]{@{}c@{}}Chimera\\ Layer 3\end{tabular}  & C2f                                                                & 48                                                                & 88                                                                 & 3                \\ \hline
\begin{tabular}[c]{@{}c@{}}Chimera\\ Layer 4\end{tabular}  & WaveMLP                                                            & 88                                                                & 120                                                                & 3                \\ \hline
\end{tabular}
\end{table}

\begin{table}[ht!]
\centering
\caption{Chimera-5M Detailed Architecture}
\footnotesize
\label{tab:chimera5m}
\begin{tabular}{|c|c|c|c|c|}
\hline
\textbf{Block}                                             & \textbf{\begin{tabular}[c]{@{}c@{}}Processing\\ Unit\end{tabular}} & \textbf{\begin{tabular}[c]{@{}c@{}}Input \\ Channel\end{tabular}} & \textbf{\begin{tabular}[c]{@{}c@{}}Output\\  Channel\end{tabular}} & \textbf{Repeats} \\ \hline
STEM                                                       & Conv2D                                                             & 10                                                                & 24                                                                 & 1                \\ \hline
\begin{tabular}[c]{@{}c@{}}Chimera \\ Layer 1\end{tabular} & MaxViT                                                             & 24                                                                & 40                                                                 & 1                \\ \hline
\begin{tabular}[c]{@{}c@{}}Chimera\\ Layer 2\end{tabular}  & C2f                                                                & 40                                                                & 72                                                                 & 3                \\ \hline
\begin{tabular}[c]{@{}c@{}}Chimera\\ Layer 3\end{tabular}  & Mamba                                                              & 72                                                                & 72                                                                 & 2                \\ \hline
\begin{tabular}[c]{@{}c@{}}Chimera\\ Layer 4\end{tabular}  & WaveMLP                                                            & 72                                                                & 152                                                                & 3                \\ \hline
\end{tabular}
\end{table}

\begin{table}[ht!]
\centering
\caption{Chimera-10M Detailed Architecture}
\footnotesize
\label{tab:Chimera10M}
\begin{tabular}{|c|c|c|c|c|}
\hline
\textbf{Block}                                             & \textbf{\begin{tabular}[c]{@{}c@{}}Processing\\ Unit\end{tabular}} & \textbf{\begin{tabular}[c]{@{}c@{}}Input \\ Channel\end{tabular}} & \textbf{\begin{tabular}[c]{@{}c@{}}Output\\  Channel\end{tabular}} & \textbf{Repeats} \\ \hline
STEM                                                       & Conv2D                                                             & 10                                                                & 24                                                                 & 1                \\ \hline
\begin{tabular}[c]{@{}c@{}}Chimera \\ Layer 1\end{tabular} & Mamba                                                              & 24                                                                & 48                                                                 & 2                \\ \hline
\begin{tabular}[c]{@{}c@{}}Chimera\\ Layer 2\end{tabular}  & C2f                                                                & 48                                                                & 88                                                                 & 3                \\ \hline
\begin{tabular}[c]{@{}c@{}}Chimera\\ Layer 3\end{tabular}  & WaveMLP                                                            & 88                                                                & 152                                                                & 3                \\ \hline
\begin{tabular}[c]{@{}c@{}}Chimera\\ Layer 4\end{tabular}  & MaxViT                                                             & 152                                                               & 192                                                                & 1                \\ \hline
\end{tabular}
\end{table}

\section{Extra Results}
\label{supp:extra}

Table \ref{tab:pedro} presents more comprehensive results for PEDRo. As a recent dataset, there is still a scarcity of related research; therefore, we have included additional models for comparative analysis. The MaxViT-Baseline was developed based on the backbone specifications from RVT-T \cite{rvt} combined with the YOLOv8 detection head from this study, adhering to the guidelines outlined in Table \ref{tab:panet_params}. This model includes 5M additional parameters compared to the original, primarily due to the design rules adopted for the Detection Heads. Its large initial channel count $Ch_{0} = 32$ resulted in a large detection head.

We also introduced a baseline model based on the WaveMLP architecture. However, since the original implementation featured models with over 30M parameters—much larger than those considered in this paper—we instantiated the WaveMLPs using the exact channel specifications as YOLOv8n. A similar approach was taken for the Mamba model. However, we did not follow the original MambaVision implementation \cite{mambavision}, as our focus was on the Mamba blocks without the ViTs. This deviation from the original template may account for the poor performance of this baseline. For the ReYOLOv8 models, we reported mAPs without the Random Polarity Suppression (RPS) augmentation, as our training setups were comparable. However, these models utilized VTEI as an encoding, and RPS was defined based on this format. Therefore, as we adopted SHIST on the Chimera architectures, we chose to compare both without this additional step, as such a technique has not yet been investigated within the current format.

\begin{table}[ht!]
\centering
\footnotesize
\caption{Comparison between the results and the state-of-the-art for the dataset PEDRo.}
\label{tab:pedro}
\begin{tabular}{|cccc|}
\hline
\textbf{Model}                                                                                       & \textbf{Network}      & \textbf{Parameters} & \textbf{mAP}  \\ \hline
\begin{tabular}[c]{@{}c@{}}Chimera-3M\\ (this work)\end{tabular}                                     & Hybrid + RNN          & 3.0M                & 59.5          \\
\begin{tabular}[c]{@{}c@{}}ReYOLOv8n\\ \cite{reyolov8}\end{tabular}                     & CNN + RNN             & 4.7M                & 59.0          \\
\textbf{\begin{tabular}[c]{@{}c@{}}Chimera-5M\\ (this work)\end{tabular}}                            & \textbf{Hybrid + RNN} & \textbf{4.9M}       & \textbf{64.2} \\ \hline
\begin{tabular}[c]{@{}c@{}}Chimera-10M\\ (this work)\end{tabular}                                    & Hybrid + RNN          & 7.2M                & 64.9          \\
\begin{tabular}[c]{@{}c@{}}WaveMLP-Baseline\\ (Based on \cite{mambavision})\end{tabular}                                        & MLP-Mixer + RNN         & 8.2M                & 59.7          \\
\begin{tabular}[c]{@{}c@{}}ReYOLOv8s\\ \cite{reyolov8}\end{tabular}                     & CNN + RNN             & 8.4M                & 64.5          \\
\begin{tabular}[c]{@{}c@{}}MaxViT-Baseline\\ (Based on \cite{rvt}) \end{tabular} & Transformer + RNN     & 9.2M                & 52.6          \\
\begin{tabular}[c]{@{}c@{}}Mamba-Baseline\\ (Based on \cite{mambavision})\end{tabular}                                          & Mamba + RNN           & 14.0M               & 46.9          \\
\textbf{\begin{tabular}[c]{@{}c@{}}ReYOLOv8m\\ \cite{reyolov8}\end{tabular}}            & \textbf{CNN + RNN}    & \textbf{18.1M}      & \textbf{66.5} \\
\begin{tabular}[c]{@{}c@{}}YOLOv8x\\ \cite{reyolov8} \end{tabular}                         & CNN                   & 68.2M               & 58.6          \\ \hline
\end{tabular}
\end{table}

\begin{figure*}[hh]
\centering
    \includegraphics[scale=0.15]{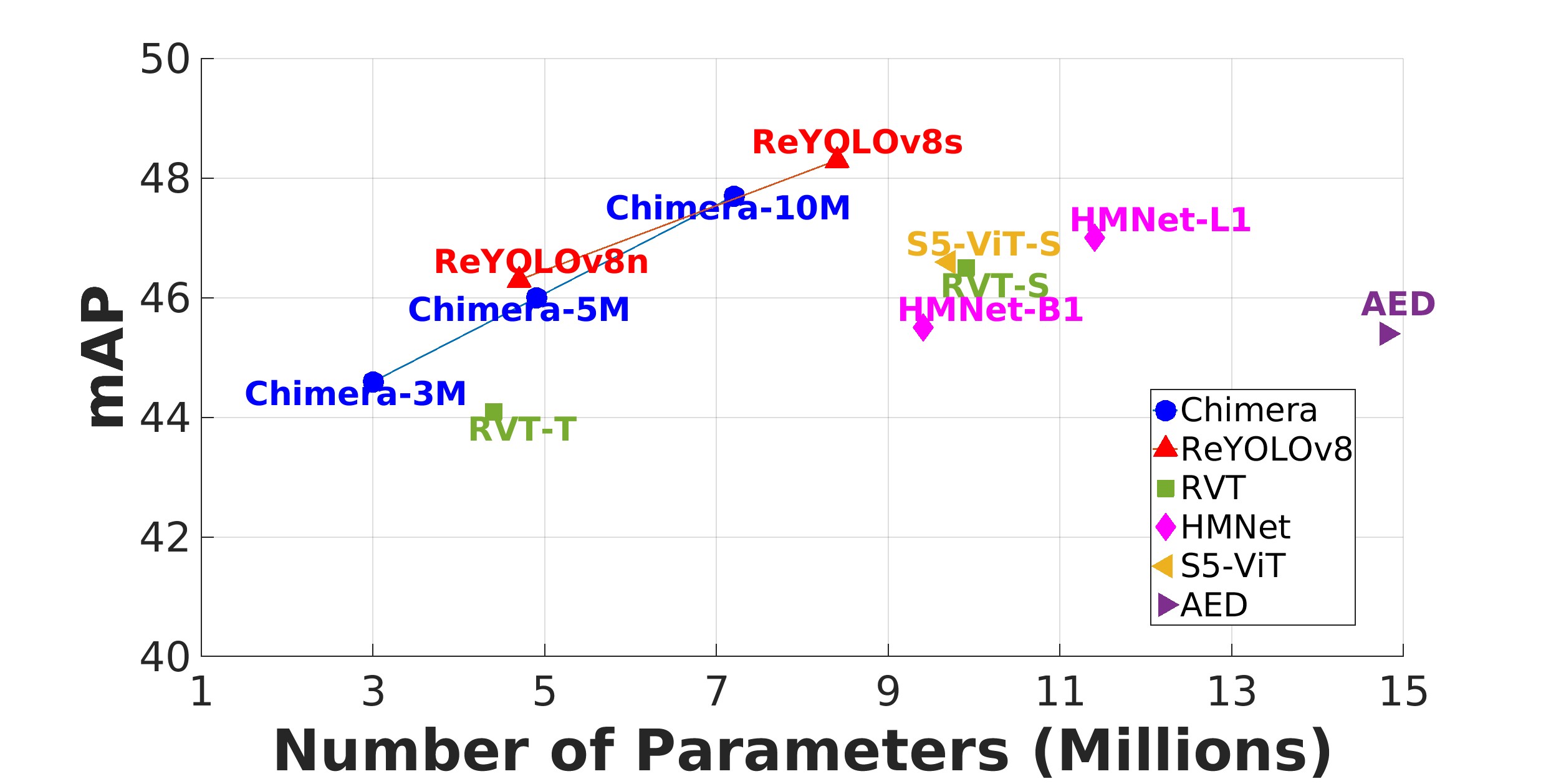}
    \caption{Comparison between the results and the state-of-the-art for the dataset GEN1.}
    \label{fig:results2}
\end{figure*}

Since the primary focus of this paper of establishing procedures for creating hybrid architectures was fully carried out through PEDRo, the results for GEN1 were reserved for this supplementary material. Table \ref{tab:gen1_sota} displays the evaluation of Chimera architectures on the GEN1 dataset compared to the literature, focusing solely on models of similar scales. As demonstrated with PEDRo, the architectures exhibit competitive performance compared to similarly-scaled models.

The first notable result is Chimera-3M, which outperformed RVT-T by 0.5 while utilizing 32\% fewer parameters. Additionally, Chimera-5M achieved nearly the same performance as ReYOLOv8n, with only a slight difference of 0.2M parameters. Another highlight is Chimera-10M, which reached 98.7\% of the mAP of ReYOLOv8s but with 14.3\% fewer parameters.

It is important to note that all optimization processes within Chimera-NAS—ranging from creating benchmarks and analyzing the ZS-NAS proxies to training the top-5 architectures in Stage 2 of Algorithm \ref{alghtm:chimera}—were conducted solely on PEDRo. Despite this, the resulting architectures demonstrated state-of-the-art performance with fewer parameters when trained from scratch on a dataset not part of the optimization process, showcasing their potential for generalization within Chimera-NAS.

\begin{table}[ht!]
\centering
\caption{Comparison with the state-of-the-art for the GEN1 dataset, up to 15M parameters}
\label{tab:gen1_sota}
\footnotesize
\begin{tabular}{cccc}
\hline
\multicolumn{1}{|c}{\multirow{2}{*}{\textbf{Model}}}                                                            & \multirow{2}{*}{\textbf{Network}} & \multirow{2}{*}{\textbf{Parameters}} & \multicolumn{1}{c|}{\multirow{2}{*}{\textbf{mAP}}} \\
\multicolumn{1}{|c}{}                                                                                           &                                   &                                      & \multicolumn{1}{c|}{}                              \\ \hline
\multicolumn{1}{|c}{\begin{tabular}[c]{@{}c@{}}Chimera-3M\\ (this work)\end{tabular}}                           & Hybrid + RNN                      & 3.0M                                 & \multicolumn{1}{c|}{44.6}                          \\
\multicolumn{1}{|c}{\begin{tabular}[c]{@{}c@{}}RVT-T \\ \cite{rvt}\end{tabular}}                   & Transformer + RNN                 & 4.4M                                 & \multicolumn{1}{c|}{44.1}                          \\
\multicolumn{1}{|c}{\textbf{\begin{tabular}[c]{@{}c@{}}ReYOLOv8n\\ \cite{reyolov8}\end{tabular}}} & \textbf{CNN + RNN}                & \textbf{4.7M}                        & \multicolumn{1}{c|}{\textbf{46.3}}                 \\
\multicolumn{1}{|c}{\begin{tabular}[c]{@{}c@{}}Chimera-5M\\ (this work)\end{tabular}}                           & Hybrid + RNN                      & 4.9M                                 & \multicolumn{1}{l|}{46.0}                          \\ \hline
\multicolumn{1}{|c}{\begin{tabular}[c]{@{}c@{}}Chimera-10M\\ (this work)\end{tabular}}                          & Hybrid + RNN                      & 7.2M                                 & \multicolumn{1}{c|}{47.7}                          \\
\multicolumn{1}{|c}{\textbf{\begin{tabular}[c]{@{}c@{}}ReYOLOv8s\\ \cite{reyolov8}\end{tabular}}} & \textbf{CNN + RNN}                & \textbf{8.4M}                        & \multicolumn{1}{c|}{\textbf{48.3}}                 \\

\multicolumn{1}{|c}{\begin{tabular}[c]{@{}c@{}}HMNet-B1\\ \cite{hmnet}\end{tabular}}               & HMNet                             & 9.4M                                 & \multicolumn{1}{c|}{45.5}                          \\
\multicolumn{1}{|c}{\begin{tabular}[c]{@{}c@{}}S5-ViT-S\\ \cite{event-ssm}\end{tabular}}           & Transformer + SSM                 & 9.7M                                 & \multicolumn{1}{c|}{46.6}                          \\
\multicolumn{1}{|c}{\begin{tabular}[c]{@{}c@{}}RVT-S \\ \cite{rvt}\end{tabular}}                   & Transformer + RNN                 & 9.9M                                 & \multicolumn{1}{c|}{46.5}                          \\
\multicolumn{1}{|c}{\begin{tabular}[c]{@{}c@{}}HMNet-L1\\ \cite{hmnet}\end{tabular}}               & HMNet                             & 11.4M                                & \multicolumn{1}{c|}{47.0}                          \\
\multicolumn{1}{|c}{\begin{tabular}[c]{@{}c@{}}AED\\ \cite{aed}\end{tabular}}                      & CNN                               & 14.8M                                & \multicolumn{1}{c|}{45.4}                          \\ \hline
\multicolumn{1}{l}{}                                                                                            & \multicolumn{1}{l}{}              & \multicolumn{1}{l}{}                 & \multicolumn{1}{l}{}                              
\end{tabular}
\end{table}

\end{document}